%% file: main.tex
\documentclass[lettersize,journal]{IEEEtran}
\usepackage{amsmath,amsfonts}
\usepackage{algorithmic}
\usepackage{algorithm}
\usepackage{array}
\usepackage[caption=false,font=normalsize,labelfont=sf,textfont=sf]{subfig}
\usepackage{textcomp}
\usepackage{stfloats}
\usepackage{url}
\usepackage{verbatim}
\usepackage{graphicx}
\usepackage{cite}
\usepackage{booktabs}

\usepackage{multirow}
\usepackage{multicol}
\usepackage{etoolbox}

\input{macro}

\begin{document}

\title{\mytitle}
\author{
\textbf{Shang-Fu Chen$^{*1}$}
\quad \textbf{Co Yong$^{*2}$}
\quad \textbf{Shao-Hua Sun$^{1,3}$}
\\
$^{1}$Graduate Institute of Communication Engineering, National Taiwan University, Taiwan
\\
$^{2}$Data Science Degree Program, National Taiwan University and Academia Sinica, Taipei, 106, Taiwan
\\
$^{3}$Department of Electrical Engineering, National Taiwan University, Taiwan
\thanks{$^{*}$Equal contribution.\\
Correspondence to: Shao-Hua Sun \texttt{<shaohuas@ntu.edu.tw>}
}
}

\markboth{IEEE Transactions on Neural Networks and Learning Systems,~Vol., No., Date}{}%



\maketitle
\IEEEpubid{%
  \begin{minipage}{\textwidth}
  \centering
  \vspace{40pt}
  \footnotesize
  © 2025 IEEE. Personal use of this material is permitted.  Permission from IEEE must be obtained for all other uses, in any current or future media, including reprinting/republishing this material for advertising or promotional purposes, creating new collective works, for resale or redistribution to servers or lists, or reuse of any copyrighted component of this work in other works..
  \end{minipage}
}

\input{text/0_abstract}
\input{text/1_intro}
\input{text/2_related}
\input{text/3_method}
\input{text/4_experiment}

\input{text/5_discussion}

\bibliographystyle{IEEEtran}
\bibliography{ref}

%

\clearpage

\input{text/appendix}

\vfill

\end{document}

%% file: macro.tex
\newcommand{\mytitle}{Restoring Noisy Demonstration for Imitation Learning with Diffusion Models}

\newcommand{\Skip}[1]{}

\newcommand{\method}{DMDR}
\newcommand{\methodFull}{Diffusion-Model-Based Demonstration Restoration}

\newcommand{\fetchpick}{\textsc{FetchPick}}
\newcommand{\fetchpush}{\textsc{FetchPush}}
\newcommand{\handrotate}{\textsc{HandRotate}}
\newcommand{\walker}{\textsc{Walker}}

\newcommand{\ie}{\textit{i}.\textit{e}.,\ }
\newcommand{\eg}{\textit{e}.\textit{g}.,\ }

\newcommand{\myfig}[1]{Figure \ref{#1}}
\newcommand{\mytable}[1]{Table \ref{#1}}
\newcommand{\myeq}[1]{Eq. \ref{#1}}
\newcommand{\mysecref}[1]{Section \ref{#1}}
\newcommand{\myalgo}[1]{Algorithm \ref{#1}}

\newcommand{\dotieconcat}[2]{
  \text{\raisebox{.8ex}{$\smallfrown$}}%
}

\makeatletter
\newcommand\dslfontsize{\@setfontsize\dslfontsize\@viipt\@viiipt}
\renewcommand\scriptsize{\@setfontsize\subfigcap{7}{8}}%
\makeatother

\usepackage{color}
\usepackage{xcolor}
\definecolor{codegray}{rgb}{0.5,0.5,0.5}

%% file: text/0_abstract.tex
\begin{abstract}
Imitation learning (IL) aims to learn a policy from expert demonstrations and has been applied to various applications. By learning from the expert policy, IL methods do not require environmental interactions or reward signals.
However, most existing imitation learning algorithms assume perfect expert demonstrations, but expert demonstrations often contain imperfections caused by errors from human experts or sensor/control system inaccuracies.
To address the above problems, this work proposes a filter-and-restore framework to best leverage expert demonstrations with inherent noise.
Our proposed method first filters clean samples from the demonstrations and then learns conditional diffusion models to recover the noisy ones.
We evaluate our proposed framework and existing methods in various domains, including robot arm manipulation, dexterous manipulation, and locomotion. 
The experiment results show that our proposed framework consistently outperforms existing methods across all the tasks.
Ablation studies further validate the effectiveness of each component and demonstrate the framework’s robustness to different noise types and levels. These results confirm the practical applicability of our framework to noisy offline demonstration data.

\end{abstract}

\begin{IEEEkeywords}
Imitation Learning, Learning from Noisy Data, Data Restoration, Behavioural Cloning, Diffusion Models
\end{IEEEkeywords}

%% file: text/1_intro.tex
\section{Introduction}
\IEEEpubidadjcol

\IEEEPARstart{I}{mitation} learning~\cite{hussein2017imitation, piot2016bridging, zheng2022imitation, zhang2022leveraging, vahabpour2024diverse, cheng2021guaranteed, ali2020review, argall2009survey, zare2024survey, chen2024diffusion, lai2024diffusion, huang2024diffusion, yeh2025action} aims to learn a policy from expert demonstrations and has been applied to various applications, including robotics{~\cite{argall2009survey}, industrial automation, strategy board games, video games, etc~\cite{fang2019survey, codevilla2018end, le2022survey, ross2010efficient, harmer2018imitation, duarte2020survey}.
Compared to reinforcement learning (RL), acquiring a policy in a trial-and-error manner, which can be unsafe or expensive, imitation learning (IL) algorithms can learn without environmental interactions.
Furthermore, while designing sophisticated RL reward functions is often difficult and tedious~\cite{amodei2016concrete, hadfield2017inverse}, IL methods learn from expert demonstrations and do not require reward signals.

Despite the wide applicability, most existing imitation learning algorithms assume perfect (\ie optimal and clean) expert demonstrations, which can be challenging and expensive to collect.
Specifically, expert demonstrations often contain imperfections caused by errors from human experts or sensor and control system inaccuracies. 
For example, the sensors may induce noises due to environmental interference~\cite{yang2000performance, biju2024assessing}, and the control system could perform imperfectly due to steady-state error or control jitter~\cite{smeds2012effect, ali2020review, gao2023intelligent}.
As a result, learning from noisy expert demonstrations using IL methods while neglecting the noises can significantly limit the performance of acquired policies~\cite{wu2019imitation, wang2021learning, zare2024survey, zheng2022imitation}.

To best leverage expert demonstrations with inherent noises, we propose to (1) filter clean demonstrations from noisy demonstrations, (2) model the clean demonstrations, (3) restore the noisy demonstrations with the learned model, and (4) aggregate the clean and restored demonstrations to learn a policy, as illustrated in~\myfig{fig:teaser}.
In contrast, most existing learning from noisy demonstration methods falls short of implementing this complete pipeline.
For example, \cite{sasaki2020behavioral, wang2021learning} filter out or give low importance weights on demonstrations determined noisy, failing to extract information from noisy demonstrations;
on the other hand, data restoration methods such as~\cite{chung2022improving, kawar2022denoising, murata2023gibbs} require a known linear degradation model, which is inaccessible for noisy demonstrations in imitation learning. 
Uniformly restoring entire demonstration sets without separating potentially clean demonstrations can incorrectly modify clean demonstrations and lead to deteriorated learning performance.

\input{fig/teaser}

This work proposes a filter-and-restore framework, \methodFull{} (\method{}), for imitation learning from noisy demonstrations.
In the demonstration filtering stage, we train autoencoders and perform the local outlier factor~\cite{breunig2000lof} using the learned embeddings to assign a pseudo label to each data point.
In the demonstration restoration stage, we consider the correlation between states and actions and train a pair of conditional diffusion models using the pseudo-labels.
One conditional diffusion model aims to restore actions based on the corresponding states, while another diffusion model focuses on the reverse, restoring states based on actions.
Then, we aggregate the clean and restored demonstrations and learn a policy using existing IL methods, such as behavioral cloning (BC)~\cite{pomerleau1989alvinn, bain1995framework}, implicit behavioral cloning (IBC)~\cite{florence2022implicit}, and diffusion policy (DP)~\cite{pearce2022imitating, chi2023diffusionpolicy}.

We evaluate our proposed framework and existing methods in various domains, including robot arm manipulation, dexterous robotic hand manipulation, and locomotion.
The experimental results show that our proposed framework consistently outperforms existing methods across all the tasks.
Also, we conduct extensive ablation studies to justify all the components in our filter-and-restore pipeline, including the filtering methods and the restoration settings.
The experimental results also confirm that our proposed filter-and-restore pipeline is IL method agnostic, \ie can be combined with various existing IL methods, including BC, IBC, and DP, and yield improved performance.
A noise-type analysis further shows that \method{} maintains considerable advantages under various noise types, including light-tailed, heavy-tailed, biased, and multi-source mixtures. The results highlight \method{}'s practical robustness to real-world demonstrations' diverse and complex noise.

%% file: fig/teaser.tex
\begin{figure}[t]
\centering
    \includegraphics[width=\linewidth]{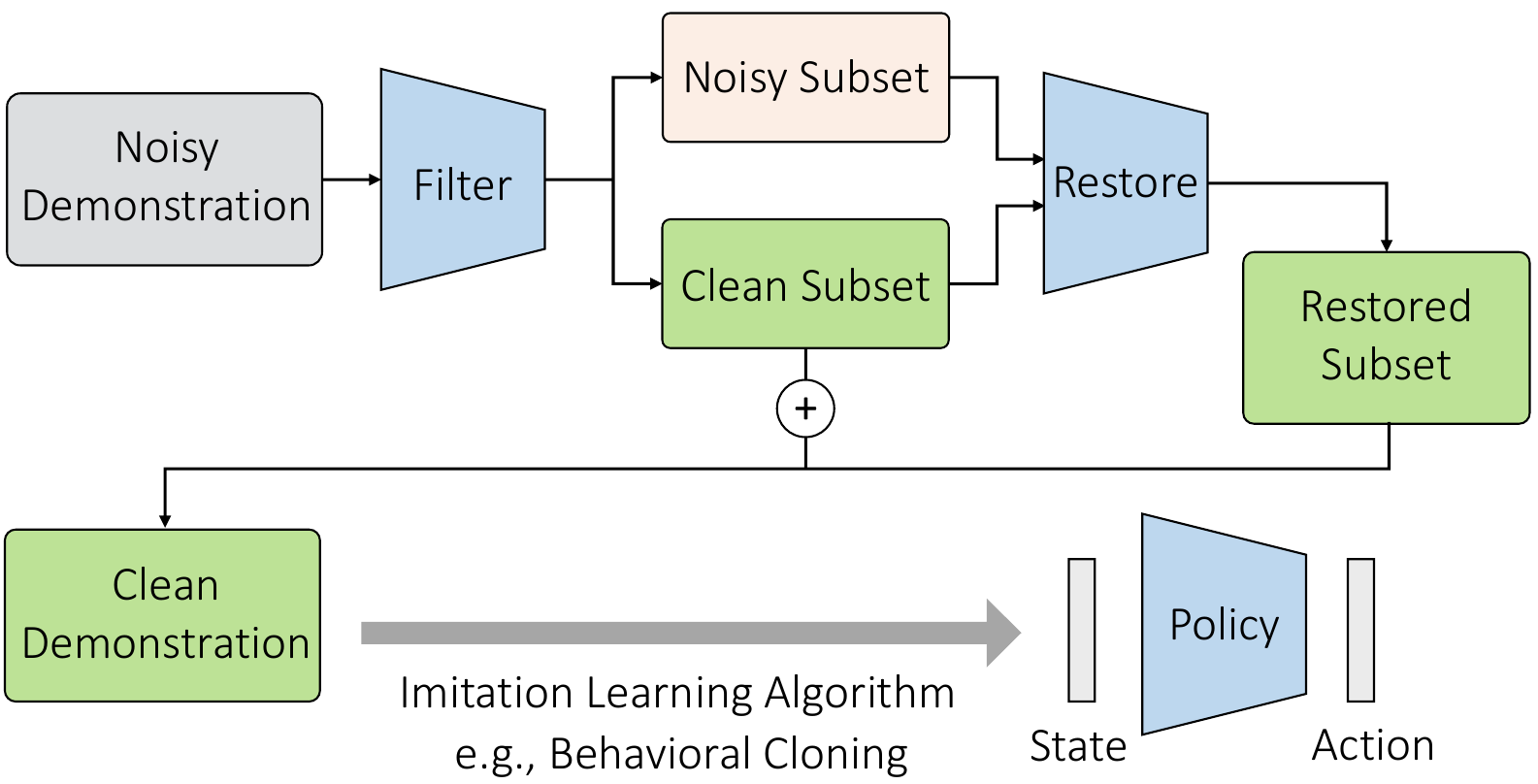} 
    \caption[]{
        \textbf{Illustration of \methodFull{} (\method{})}: 
        We propose a two-stage learning framework that first identifies and filters clean samples from the noisy demonstrations. Then, by learning diffusion models using the clean samples, we restore the remaining noisy samples to provide more reliable demonstrations.
        \label{fig:teaser}
    }
\end{figure}

%% file: text/2_related.tex
\section{Related Works}
\subsection{Imitation Learning (IL) from expert demonstrations}
Imitation learning aims to learn a policy by observing expert demonstrations without reward signals from the environment. 
Online imitation learning methods use rollouts collected from online interaction to help policy learning.
For instance, generative adversarial imitation learning (GAIL) methods~\cite{ho2016generative, torabi2018generative} learn discriminators that can identify the rollout samples and the expert samples for training a policy that models the expert distribution.
Closely related to imitation learning, inverse reinforcement learning (IRL) methods~\cite{fu2018learning, lee2021generalizable, piot2016bridging} aim to derive the reward function from expert demonstrations for policy learning; Reinforcement Learning from Expert Demonstrations (RLED)~\cite{jing2020reinforcement, liu2022improved, ramirez2022model} is a paradigm in reinforcement learning that combines expert demonstrations with traditional reinforcement learning techniques to improve learning efficiency and performance. It leverages expert trajectories—sequences of states, actions, and sometimes rewards—to guide the agent's exploration and policy learning.

On the other hand, offline imitation learning methods learn a policy directly from a fixed set of expert demonstrations without environmental interactions.
These methods are beneficial when online interactions are expensive, risky, or impractical.
Behavior cloning (BC)~\cite{michie1990cognitive, bain1995framework, ijcai2018p687} is a widely studied offline learning approach aimed at imitating expert behaviors through supervised learning;
Implicit BC (IBC)~\cite{florence2022implicit} learns an energy-based model that takes both states and actions as inputs for better generalization ability;
Diffusion Policy~\cite{pearce2022imitating, chi2023diffusionpolicy} employs a diffusion model as a policy to capture multi-modal behaviors that BC struggles to model. 
However, the effectiveness of these imitation learning methods heavily depends on the quality of the expert demonstrations. Noisy samples in demonstrations can significantly hinder the learning process and the performance of the derived policies~\cite{zare2024survey, zheng2022imitation, Xiao_2025}.

\subsection{Imitation Learning (IL) from Noisy Demonstrations}
The issue of noisy demonstrations poses unique challenges to policy learning since the polluted data or trajectories could severely degrade the performance of policy learning.
Similar to the case of IL from clean expert demonstrations, recent works have explored online and offline approaches with distinct designs.

For online IL from noisy demonstrations, ~\cite{brown2019extrapolating, brown2020better} proposed methods that aim to derive a reward function from suboptimal demonstrations in order to extrapolate better trajectories during inference.
Generative adversarial-based approaches~\cite{wu2019imitation, Tangkaratt2020RobustIL} assign a confidence or optimality score for training samples to alleviate the interference of the noises.
However, in order to evaluate the noisy demonstrations, these online IL methods usually leverage a supplementary dataset with confidence annotations, but such clean and well-annotated data samples are usually hard to collect in real-world cases.

Offline IL methods, which do not require environmental interactions, rely more on the quality of the provided expert demonstrations, making it more challenging to learn effective policies from noisy demonstrations.
\cite{kim2021demodice} proposes learning an estimation of the stationary distribution to regularize policy learning and alleviate the disturbance of imperfect demonstrations.
~\cite{yu2023offline} extends this approach by imposing additional constraints on optimization to consider the diversity of both states and actions.

Discriminator-based methods~\cite{xu2022discriminator, zhang2023discriminator} assign weights for training samples based on their suboptimality using discriminators in order to emphasize the effect of good data samples while ignoring the bad ones.
Behavioral Cloning from Noisy Demonstrations (BCND)~\cite{sasaki2020behavioral} assigns weights for training samples based on predictions from previous iterations to seek the major mode of the distribution.
However, these methods give low importance weights on demonstrations determined as noisy and fall short of extracting useful information from them.

To address the above issues, we propose a filter-and-restore framework that restores noisy demonstrations to best leverage the noisy demonstrations for policy learning. In our experiments, we compare our proposed framework with BCND to demonstrate the robustness and effectiveness of our approach.

\subsection{Anomaly Detection (Outlier Detection)} 
Anomaly detection~\cite{li2021automated} aims to identify samples that deviate from the expected, normal, or typical patterns. 
Similarly, expert demonstrations represent the \textit{normal} or expected behavior in the domain of imitation learning.
Accordingly, we consider corrupted or irrelevant parts of demonstrations as anomalies, such as those caused by sensor glitches, actuator failures, or off-task behaviors.

From the aspect of data characteristics, abnormal samples often stem from rare occurrences or unexpected events, making them inherently diverse and challenging to collect~\cite{Zhu2025_StructAttkGAD}.
Therefore, previous literature focuses on learning models to identify the distribution of normality and classify data that deviates from the learned distributions as anomalies.

Classification-based anomaly detection~\cite{ruff2018deep, ruff2019deep, wang2019semi, hendrycks2018deep, sohn2020learning, li2021cutpaste} formulates anomaly detection as a one-class classification problem, where only normal samples are available during training.
For instance,~\cite{ruff2019deep, wang2019semi} utilize a limited set of labeled outliers samples with unlabeled samples to train the classifiers.
Alternatively, other works~\cite{hendrycks2018deep, sohn2020learning, li2021cutpaste} augment the training data with out-of-distribution or synthetic samples to enhance the classifier's ability to recognize anomalies. Nevertheless, applying these augmentation methods to offline imitation learning is challenging due to the significant differences in behavior among different environments. 

On the other hand, reconstruction-based anomaly detection methods utilize autoencoders~\cite{zong2018deep, gong2019memorizing, park2020learning, tan2021trustmae, Kong2025_FedGraphAD, Huang2024_ScoreNetAD} to determine outliers based on the reconstruction errors. 
In these approaches, neural networks are trained to reconstruct input data, with a widely used assumption that deep models tend to learn clean samples faster than noisy samples~\cite{ren2018learning, li2020dividemix, karim2022unicon}.
Consequently, samples with higher reconstruction errors are often flagged as anomalies.
In this work, we apply a reconstruction-based method to filter outliers for noisy demonstrations

\subsection{Diffusion Models for Data Restoration} Diffusion models have demonstrated remarkable performance in various generation tasks~\cite{rombach2022high, ho2022video, ho2020denoising, song2021denoising, song2023consistency, wang2023diffusion, moser2024diffusion, zhao2024diffusionvmr, chen2024g, huang2024learning, lee2024dmesh}.
Recent works~\cite{kawar2022denoising, murata2023gibbs, Pei2024_TrafficImputeDiff} study to extend diffusion models for data restoration tasks, including super-resolution, inpainting, and colorization, etc. 

Denoising Diffusion Restoration Models (DDRM)~\cite{kawar2022denoising} demonstrate that when the restoration task can be formulated as a linear inverse problem, pre-trained diffusion models can be effectively leveraged for inference given the degradation matrix. This approach allows for the utilization of existing models without the need for retraining or fine-tuning for specific restoration tasks. 
Building upon DDRM, GibbsDDRM~\cite{murata2023gibbs} relaxes the requirement of the pre-defined degradation matrix and adopts a learnable linear operator to describe the restoration task instead.

However, existing works in this domain often rely on diffusion models pre-trained on clean datasets, which may not be directly applicable to noisy demonstrations commonly encountered in real-world scenarios. To address this limitation, we propose a two-stage framework. In the first stage, we filter clean samples from noisy demonstrations to create a dataset suitable for training conditional diffusion models. In the second stage, we utilize the derived diffusion models to restore the noisy data effectively.

%% file: text/3_method.tex
\section{Method}

In this paper, we aim to learn from noisy demonstrations without environmental interactions or additional annotations for clean samples.
The noisy demonstrations $D={\tau_{1},...,\tau_{M}}$ consists of $M$ trajectories, where each trajectory $\tau_i$ comprises a sequence of $n_i$ state-action pairs ${s^i_1, a^i_1, ..., s^i_{n_i}, a^i_{n_i}}$. 
For each state and action in the demonstration, there exists a small probability $p$ that random noise is injected, indicating the noise level. 
Notably, the pollution of states and actions is independent since distinct sensors are typically used for monitoring states and actions in real-world scenarios. 
To simplify notation, we denote a state-action pair sampled from the entire demonstration dataset as $(s, a) \sim D$, dismissing the trajectory index and the index of a state-action pair within a trajectory.

We propose a \emph{filter--and--restore} framework for imitation learning from noisy demonstrations.  
Given the noisy demonstrations $D$, the \textbf{Demonstration Filtering} stage (\mysecref{sec:demo_filter}) combines an autoencoder with Local Outlier Factor to flag anomalous elements, denoted as $s'$ or $a'$. In contrast, those who pass the test are marked as $\hat{s}$ or $\hat{a}$, and treated as \emph{clean}.  
The \textbf{Demonstration Restoration} stage (\mysecref{sec:demo_restore}) trains conditional diffusion models on these clean samples and then maps each flagged component $s'$ or $a'$ to its denoised counterpart $s^{*}$ or $a^{*}$, resulting in \emph{restored} pairs.  
After these two steps, we obtain a refined dataset composed of fully clean pairs $(\hat{s},\hat{a})$ together with partially or fully restored pairs $(s^{*},\hat{a})$, $(\hat{s},a^{*})$, and $(s^{*},a^{*})$, which serves as reliable supervision for downstream policy learning.

\subsection{Demonstration Filtering}
\label{sec:demo_filter}
\input{fig/filter}
In this stage, we aim to filter clean samples from noisy demonstrations by integrating autoencoders and Local Outlier Factor (LOF).
The process involves learning global features of the data distribution with autoencoders and then applying LOF to identify potential outliers based on local densities.
While demonstrations typically consist of sequences of state-action pairs, in many applications, states and actions represent distinct properties and are usually captured by different sensors. 
For instance, states commonly denote variables like position or RGB images, whereas actions typically record the torque applied to the joints of robots, as seen in the widely utilized MuJoCo environment~\cite{todorov2012mujoco}, Pybullet~\cite{coumans2021}, OpenAI Gym~\cite{1606.01540}, Isaac Gym~\cite{makoviychuk2021isaac}, etc.
The noisy demonstrations may arise from control failures that occurred during the data collection process.
For instance, performing the desired actions (applying desired torques on joints) could face control system errors, such as steady-state error and control jitter~\cite{smeds2012effect, ali2020review, gao2023intelligent}, so states and actions would encounter noise perturbation independently.
It is worth noting that the goal of this paper is to learn a policy from given noisy demonstrations instead of solving internal control problems directly.
As a result, we filter and label state and action samples independently.

\subsubsection{Anomaly Detection with Autoencoders}

We apply a reconstruction-based method to detect and filter potential abnormal samples from noisy demonstrations.
As depicted in~\myfig{fig:filter_training}, we train a pair of autoencoders to capture the majority of states and actions with a reconstruction loss, which can be formulated as follows:
\begin{equation}
\begin{aligned}
\label{eq:rec_loss_s} 
    \mathcal{L}^{s}_{\text{rec}} = \mathbb{E}_{(s, a) \sim D}\left[{||\phi(s) - s||}^2\right],
\end{aligned}
\end{equation}
and
\begin{equation}
\begin{aligned}
\label{eq:rec_loss_a} 
    \mathcal{L}^{a}_{\text{rec}} = \mathbb{E}_{(s, a) \sim D}\left[{||\phi(a) - a||}^2\right],
\end{aligned}
\end{equation}
where $\phi(s)$ indicates the state autoencoder and $\phi(a)$ indicates the action autoencoder.
One can directly adopt $L^{s}_{rec}$ and $L^{a}_{rec}$ to filter outliers as reconstruction-based anomaly detection approaches do.

However, in tasks like robot arm manipulation, we find that certain state-action pairs may be infrequent yet crucial for successful execution. 
For instance, in a scenario where a robot arm needs to grasp an object and arise it to a target location. 
While most state-action pairs may correspond to routine movements, such as navigating the arm, there are specific instances, such as the action of grasping an object, that occur less frequently but are indispensable for task completion.
To prevent discarding essential samples, we further apply the Local Outlier Factor algorithm to identify outliers based on the encoded representations obtained from autoencoders.

\subsubsection{Combining Autoencoder and Local Outlier Factor}

Local Outlier Factor (LOF) is known as an anomaly detection algorithm that measures the local deviation of samples.
The algorithm begins by estimating the local density for each sample in the dataset using the $k$-nearest neighbors (KNN) approach. It calculates the density by considering the distance to each sample's $k$ nearest neighbors.
Next, a ratio of local density is computed using the sample and its neighbors, which serves as the LOF score.
A larger LOF score indicates that the sample has a lower density than its neighbors, suggesting it may be an outlier.
LOF effectively identifies potential outliers in the dataset by analyzing each local region individually.

As we mentioned in the previous section, to prevent predicting state-action pairs with unique behaviors as anomalies, we utilize the bottleneck features $z_s$ and $z_a$ from the autoencoders to calculate LOF score for each $s$ and $a$.
These representations capture global behaviors since the autoencoders are trained to minimize reconstruction loss across the entire expert demonstration dataset. 
LOF then evaluates outliers based on the local density of these representations.
By considering the local density of samples' representations, we effectively prevent discarding infrequent samples with useful behaviors, thereby enhancing the robustness of demonstration filtering for imitation learning tasks.

The goal of anomaly detection is to identify samples that differ from normal or typical patterns. 
Therefore, in anomaly detection literature~\cite{breunig2000lof, liu2008isolation, ruff2018deep}, data that deviate from the majority are typically regarded as anomalies. 
This assumes that the majority of the training dataset, meaning at least half of the samples, is composed of normal data.
Following this assumption, half of the samples with lower LOF scores are labeled as clean ($\hat{s}$ and $\hat{a}$), while the other half is labeled as noisy ($s'$ and $a'$). This design choice will be further discussed in~\mysecref{sec:TrustedFraction}.
As shown in~\myfig{fig:filter_inference}, we filter the dataset into four subsets according to the pseudo-labels: clean state-action pairs $D_{(\hat{s}, \hat{a})}$, clean states with noisy actions $D_{(\hat{s}, a')}$, noisy states with clean actions $D_{(s', \hat{a})}$, and noisy state-action pairs $D_{(s', a')}$.

\subsection{Demonstration Restoration}
\label{sec:demo_restore}

In this section, we show how to restore the noisy subsets of demonstrations with diffusion models.
Previous works~\cite{kawar2022denoising, murata2023gibbs} have shown how to use diffusion models for image restorations when the distortion process is known.
For instance, Denoising Diffusion Restoration Models (DDRM)~\cite{kawar2022denoising} utilizes a given degradation matrix for each restoration task to solve the linear inverse problem and GibbsDDRM~\cite{murata2023gibbs} assumes the distortion can be modeled by a learnable blurring kernel.
However, such information is not available in offline imitation learning.  
To restore demonstrations without a given degradation matrix, we utilize conditional diffusion models to consider the relationships between the corresponding state and action.
Moreover, we introduce noise timestep predictors to guide the denoising process of the diffusion models for accurate restoration.
We elaborate on the learning of these components in the following subsections.

\input{fig/restore_training}

\subsubsection{Denoising Diffusion Probabilistic Models (DDPMs)}
This work uses DDPMs~\cite{ho2020denoising} for data restoration. 
During the training stage, DDPMs gradually add Gaussian noise to each data sample until it becomes isotropic Gaussian, which is called the forward diffusion process.
Then, DDPMs learn to denoise the noise-injected sample to the original data, called the reverse diffusion process.
Given a data point $x_0$ sampled from dataset $D$, \eg a state $s$ or an action $a$ sampled from the demonstrations, latent variables $x_1, ..., x_T$ are produced in the forward diffusion process, where $T$ is the number of diffusion steps, and $x_T$ is an isotropic Gaussian.
The diffusion model $\theta$ learns to reverse the diffusion process by predicting the injected noise $\epsilon$ on the sample.
The objective of DDPM can be formulated as the following:
\begin{equation}
\begin{aligned}
\label{eq:diff_loss} 
    \mathcal{L}_{\text{diff}} = \mathbb{E}_{t \sim \mathcal{U}(0, T), x \sim D}\left[{||\epsilon - \epsilon_{\theta}(\alpha_{t}x_0 + \sigma_{t}\epsilon, t)||}^2\right],
\end{aligned}
\end{equation}
where $t$ is sampled from a uniform distribution $\mathcal{U}(0, T)$ and
$\sigma_{t} = \sqrt{1-{\alpha_{t}}^2}$ is a scalar representing increasing noise schedule.
Once the diffusion model $\theta$ is learned, one can sample a random noise and use the predicted noise $\epsilon_{\theta}$ to compute the next latent variable. The above process is repeated iteratively until a clean sample $x_0$ is generated.

The reverse process of DDPMs, \ie step-by-step denoising by predicting the added noise, is aligned with the nature of the signal restoration.
From the theoretical aspects, \cite{ho2020denoising, nichol2021improved} demonstrate that the diffusion model is able to optimize the variational lower-bound of the log-likelihood of the data distribution $p(x_0)$, and outperform other likelihood-based models, such as autoregressive models and variational autoencoders (VAEs);
\cite{song2020score} offers an alternative perspective by formulating the diffusion process as a continuous-time stochastic differential equation (SDE) in the limit as $T \to \infty$. This interpretation transforms the problem into denoising score matching, ensuring that the trained model converges to an optimal denoiser.
The above evidence motivates us to apply DDPMs to restore expert demonstrations polluted by noise. 

\subsubsection{Learning Conditional Diffusion Models}
To consider the correlation between states and actions, we train a pair of conditional DDPMs for states and actions, respectively, using the filtered subset containing clean state-action pairs $D_{(\hat{s}, \hat{a})}$.
The state diffusion model $\theta_{s}$ aims to restore states based on the corresponding actions, while the action diffusion model $\theta_{a}$ focuses on restoring actions based on states.
The state diffusion model $\theta_{s}$ considers the corresponding action to predict the noise-injected states.
The objective can be calculated as follows:
\begin{equation}
\begin{aligned}
\label{eq:diff_state_loss} 
\mathcal{L}^{s}_{\text{diff}} = 
\mathbb{E}_{t \sim \mathcal{U}(0, T),  (\hat{s}, \hat{a}) \sim D_{(\hat{s}, \hat{a})}}\left[{||\epsilon - \epsilon_{\theta_s}(s_t, \hat{a}, t)||}^2\right],
\end{aligned}
\end{equation}
where $s_t = \alpha_{t}\hat{s} + \sigma_{t}\epsilon$ is the noise-injected state, $\hat{a}$ is the corresponding clean action, and t is the noise timestep sampled from a uniform distribution over diffusion steps.
Similarly, we can define the objective for the action diffusion model $\theta_{a}$ as follows:

\begin{equation}
\begin{aligned}
\label{eq:diff_action_loss}
\mathcal{L}^{a}_{\text{diff}} = 
\mathbb{E}_{t \sim \mathcal{U}(0, T),  (\hat{s}, \hat{a}) \sim D_{(\hat{s}, \hat{a})}}\left[{||\epsilon - \epsilon_{\theta_a}(a_t, \hat{s}, t)||}^2\right],
\end{aligned}
\end{equation}
where $a_t = \alpha_{t}\hat{a} + \sigma_{t}\epsilon$ is the noise-injected action, $\hat{s}$ is the corresponding clean state, and $t$ is the sampled noise timestep.

Both conditional diffusion models are trained on the subset of clean pairs $D_{(\hat{s}, \hat{a})}$ to ensure the learned models can accurately capture the correct relationships between the states and actions.
After learning the diffusion models, we can apply the state diffusion model $\theta_{s}$ on the subset $D_{(s', \hat{a})}$ to restore the noisy states conditioned on the corresponding clean action.
Also, the action diffusion model $\theta_{a}$ is applied on $D_{(\hat{s}, a')}$ to restore the noisy actions.
We discard the noisy state-action pairs from the subset $D_{(s', a')}$ since the polluted state and action can not provide sufficient information for restoration.

\subsubsection{Learning Noise Timestep Predictors}
\label{sec:method_predictor}
In the previous section, we have derived diffusion models that can gradually denoise a sampled isotropic noise to a clean state or action with the iterative reverse diffusion process.
However, the noisy states and actions are still more informative than isotropic Gaussian noises despite being polluted by noise.
To best leverage these samples, we aim to predict a noise timestep $t$ for each noisy sample during the denoising process.
With the predicted noise timesteps, the trained diffusion models are able to perform restoration according to the scale of noise on each sample.
To this end, we introduce a pair of noise predictors $\psi_{s}$ and $\psi_{a}$ to predict the noise timestep for states and actions, respectively.

Similar to the diffusion models, the noise timestep predictors are conditioned on the corresponding state or action.
The training objective for the state noise predictor $\psi_{s}$ is to predict the correct noise timestep for noise-injected states from the clean subset $D_{(\hat{s}, \hat{a})}$, which can be calculated as follows:
\begin{equation}
\begin{aligned}
\label{eq:pred_state_loss} 
\mathcal{L}^{s}_{\text{pred}} = 
\mathbb{E}_{t \sim \mathcal{U}(0, T),  (\hat{s}, \hat{a}) \sim D_{(\hat{s}, \hat{a})}}\left[{||t - \psi_{s}(s_t, \hat{a})||}^2\right],
\end{aligned}
\end{equation}
where $s_t = \alpha_{t}\hat{s} + \sigma_{t}\epsilon$ is the noise-injected state, and $\hat{a}$ is the clean action to be served as the condition.
Similarly, we can learn the action noise predictor $\psi_{a}$ by the following equation: 
\begin{equation}
\begin{aligned}
\label{eq:pred_action_loss} 
\mathcal{L}^{a}_{\text{pred}} = 
\mathbb{E}_{t \sim \mathcal{U}(0, T),  (\hat{s}, \hat{a}) \sim D_{(\hat{s}, \hat{a})}}\left[{||t - \psi_{a}(a_t, \hat{s})||}^2\right],
\end{aligned}
\end{equation}
where $a_t = \alpha_{t}\hat{a} + \sigma_{t}\epsilon$ is the noise-injected action, and $\hat{s}$ is the clean state to be served as the condition.

\input{fig/restore_inference}
\subsubsection{Restoration with Diffusion Models and Noise Predictors}
The restoration of noisy states in $D_{(s', \hat{a})}$ and noisy actions in $D_{(\hat{s}, a')}$ follows analogous approaches.
Here, we illustrate how to restore the noisy states in $D_{(s', \hat{a})}$ with the trained state noise predictor $\psi_{s}$ and the state diffusion model $\theta_{s}$.

As shown in~\myfig{fig:restore_inference}, we sample a state-action pair from $D_{(s', \hat{a})}$, where the state $s'$ is noisy, and the action $\hat{a}$ is clean, predicted in the previous filtering stage. 
To restore the noisy state, we first estimate the noise timestep $t^{*}$, which measures the deviation from the major behaviors in the noisy demonstrations.
Since we only keep half of the samples in the clean subset during the filtering stage, it is possible that there are clean samples remaining in the noisy subsets as well.
Therefore, we use $t^{*}$ to filter samples by applying a noise timestep threshold $t_{\text{thres}}$. 
If $t^{*} < t_{\text{thres}}$, the state is assumed clean and added directly to the clean dataset $D_{(\hat{s}, \hat{a})}$. Otherwise, the state $s'$ is restored using $\hat{a}$ and $t^{*}$ via the state diffusion model, and the restored pair is added to the clean dataset.
The noisy actions in $D_{(\hat{s}, a')}$ can be restored following a similar procedure.

The effectiveness of the predictors and the noise timestep thresholding are studied in the~\mysecref{sec:exp_restore}.
While treating all noisy samples as isotropic Gaussian is a viable approach for restoration, our experiments demonstrate that incorporating predictors and noise timestep thresholding enhances the accuracy of diffusion models in restoration tasks.

%% file: fig/filter.tex
\begin{figure*}[!t]
    \centering  
    \subfloat[][Training of Demonstration Filtering]{\includegraphics[width=0.4\textwidth]{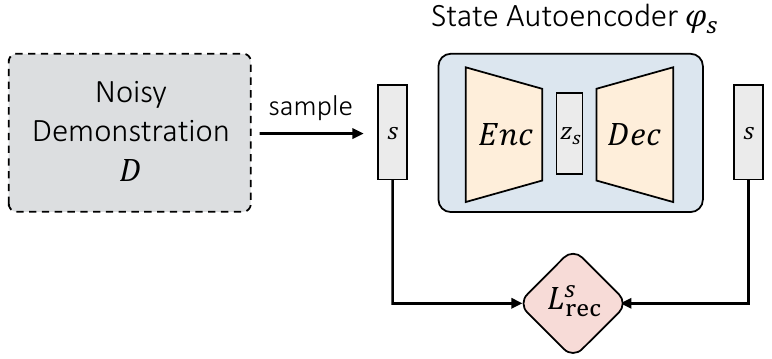}%
    \label{fig:filter_training}}
    \hspace{0.3cm}
    \subfloat[Inference of Demonstration Filtering]{\includegraphics[width=0.57\textwidth]{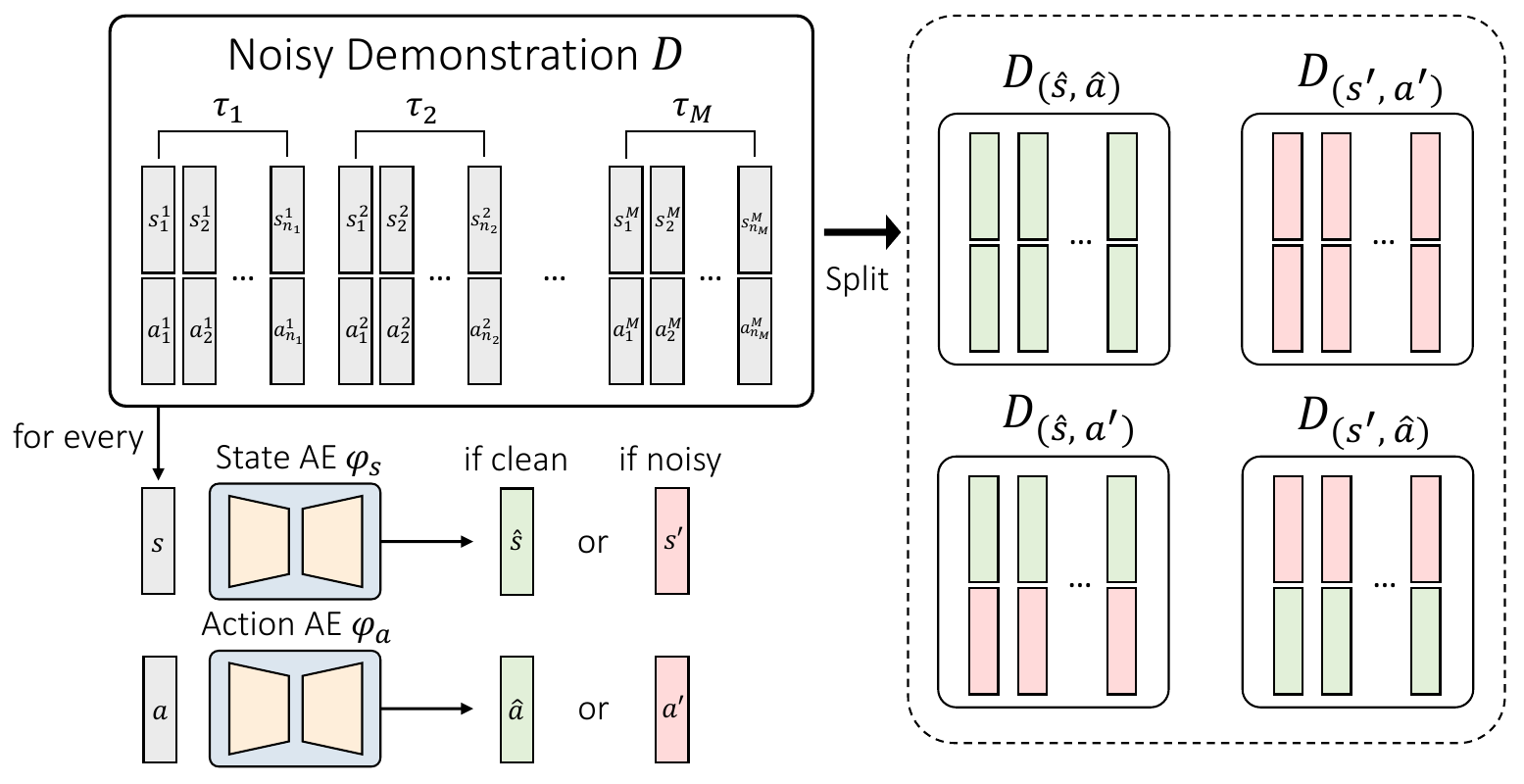}%
    \label{fig:filter_inference}}
    \caption[]{
    \textbf{Demonstration filtering.}  
    \textbf{(a) Training.} We train the state autoencoder ${\phi}_s$ using the reconstruction loss and apply the Local Outlier Factor (LOF) on the feature space, \ie $z_s$.
    We only show the state autoencoder ${\phi}_s$ in the figure for illustration, while the action autoencoder ${\phi}_a$ follows the identical architecture.
    \textbf{(b) Inference.} We use autoencoders and LOF to individually identify outliers for states and actions. With the predictions of outliers, we filter the noisy demonstrations into four subsets: $D_{(\hat{s}, \hat{a})}$, $D_{(s', a')}$, $D_{(\hat{s}, a')}$, and $D_{(s', \hat{a})}$, which contains clean state-action pairs, noisy state-action pairs, clean states with noisy actions, and noisy states with clean actions, respectively.
    }
    \label{fig:filter}
    \vspace{-3mm}
\end{figure*}

%% file: fig/restore_training.tex
\begin{figure*}[!t]
    \centering  
    \subfloat[][Training of Conditional Diffusion Models]{\includegraphics[width=0.37\textwidth]{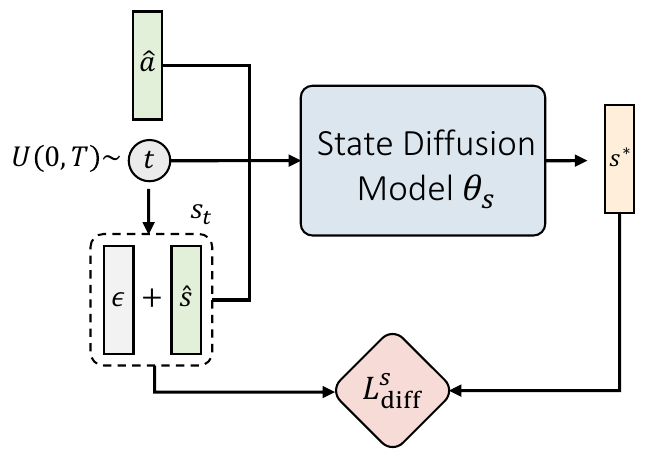}%
    \label{fig:diffusion_training}}
    \hspace{1cm}
    \subfloat[Training of Noise Timestep Predictors]{\includegraphics[width=0.45\textwidth]{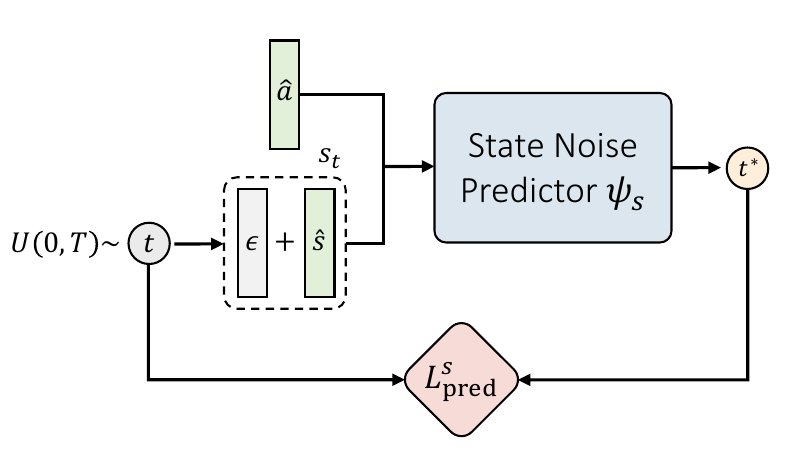}%
    \label{fig:predictor_train}}
    
    \caption[]{
    \textbf{Demonstration restoration.}  
    \textbf{(a) Training of conditional Diffusion Models.} We train the state diffusion model $\theta_s$ using the condition of clean action $\hat{a}$ to restore the noise-injected state $s_t$. We employ an identical architecture for the action diffusion model $\theta_a$, which restores the noise-injected action $a_t$ using the clean state $\hat{s}$. 
    \textbf{(b) Training of Noise timestep predictors.} To estimate the appropriate noise timestep during inference, we train the state noise predictor $\psi_{s}$ given a clean action $\hat{a}$. The action noise predictor $\psi_{a}$ follows the identical architecture.
    }
    \label{fig:restore_training}
\end{figure*}

%% file: fig/restore_inference.tex
\begin{figure}[t]
\centering
    \includegraphics[width=\linewidth]{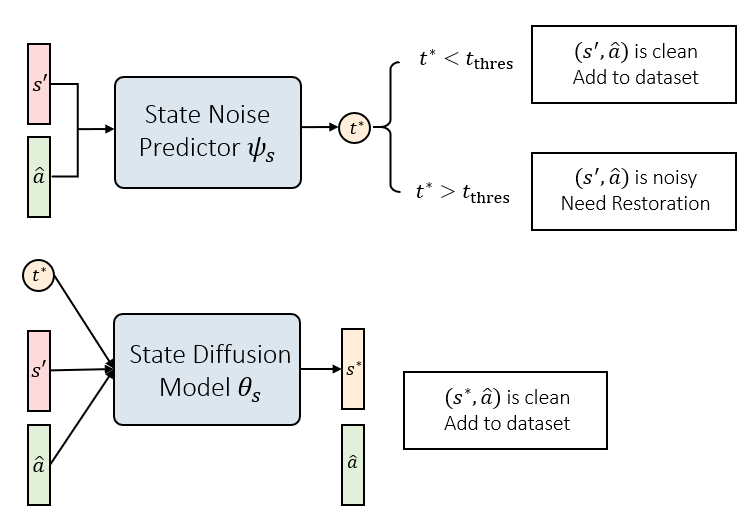} 
    \caption[]{
        \textbf{Inference of demonstration restoration.} Given a noisy state $s'$ and a clean action $\hat{a}$ from $D_{(s', \hat{a})}$, we first predict the noise timestep $t^{*}$ for the noisy state.
        If $t^{*}$ is less than a predefined threshold $t_{\text{thres}}$, then we append it to the clean subset $D_{(\hat{s}, \hat{a})}$ directly.
        Otherwise, we denoise the noisy state using the state diffusion model $\theta_{s}$, conditioned on the clean action and the predicted noise timestep.
        Similarly, noisy actions in $D_{(\hat{s}, a')}$ can be restored using the action noise predictor $\psi_{a}$ and the action diffusion model $\theta_{a}$.
    }
    \label{fig:restore_inference}
\end{figure}

%% file: text/4_experiment.tex
\section{Experiments}
\label{sec:exp}

In this section, we evaluate how our \methodFull{} (\method{}) benefits offline imitation learning with noisy demonstrations.  
We first introduce the environmental setup for the experiments (\mysecref{sec:exp_setup}) and the baselines (\mysecref{sec:exp_baseline}).
The experimental results (\mysecref{sec:exp_results}) show that \method{} is more effective than other baselines on various continuous control tasks. 
We then do the ablation studies to verify our design choice for both the filtering stage (\mysecref{sec:exp_filter} and \mysecref{sec:TrustedFraction}) and the restoration stage (\mysecref{sec:exp_restore}).
Finally, Sections~\ref{sec:exp_il_alg} to \ref{sec:noise_level}
establish the generalizability and robustness of \method{} by showing consistent performance gains between imitation learning algorithms, diverse types of noise, and different levels of noise.

\subsection{Experimental Setup}
\label{sec:exp_setup}
This paper addresses offline imitation learning using noisy demonstrations \( D \), which consist of sequences of unlabeled \((s, a)\). Since state measurements (\(s\)) and action controls (\(a\)) are typically derived from different sensors and actuators, we assume they are polluted by independent noise.

To simulate noisy demonstrations, we add noise to expert data \((s_e, a_e)\) as follows:  

\begin{equation}
\begin{aligned}
\label{eq:state_noise} 
s =  
\begin{cases}  
s_e + n_s, & \text{with probability } p, \\  
s_e, & \text{with probability } 1 - p,  
\end{cases}    
\end{aligned}
\end{equation}

\begin{equation}
\begin{aligned}
\label{eq:action_noise}   
a =  
\begin{cases}  
a_e + n_a, & \text{with probability } p, \\  
a_e, & \text{with probability } 1 - p,  
\end{cases}   
\end{aligned}
\end{equation}  

Here, noise level \(p\) denotes the probability of a state or action being corrupted by noise. 

Following previous studies that identify zero-mean Gaussian as the dominant sensor noise~\cite{gabela2019effect,el2020blind,biju2024assessing}, we inject
$n_s \sim \mathcal{N}(0,\sigma_s^{2})$ and
$n_a \sim \mathcal{N}(0,\sigma_a^{2})$ in the main experiments.  
Other noise distributions (Uniform, Laplacian, biased, and mixed variants) are analyzed separately in~\mysecref{sec:noise_type}.

All expert data ($s_e$, $a_e$) used in our experiments are generated using pre-trained reinforcement learning agents, specifically using Proximal Policy Optimization (PPO) and Soft Actor-Critic (SAC) algorithms. These agents are optimized using reward functions provided by the simulation environments, ensuring that the demonstrations reflect expert-level performance. 
For the simulation environments, we utilized the MuJoCo physics engine~\cite{todorov2012mujoco} and the OpenAI Gym~\cite{1606.01540} toolkit for our simulations. MuJoCo (Multi-Joint dynamics with Contact) is a physics engine designed for research and development in robotics, biomechanics, graphics, and animation; OpenAI Gym is a toolkit for developing and comparing reinforcement learning algorithms. It provides a variety of environments, including classic control tasks, Atari games, and robotic simulations.
Details of model architectures and hyperparameters can be found in our supplementary materials in~\mysecref{sec:app_details}.

We employ the proposed framework (\method{}) in various continuous control domains.
As shown in~\myfig{fig:environment}, we illustrate the environments and tasks used in the experiments in the following:

\input{fig/env}
\begin{itemize}
    \item \textbf{Robot Arm Manipulation.}
    We leverage the \fetchpick{} and \fetchpush{} environments to represent the robot arm manipulation tasks.
    These environments are designed for the Fetch Mobile Manipulator, ensuring that state and action representations align with the real-world robot system to enhance potential real-world applicability.
    These tasks aim to control a 7-DoF robot arm to interact with an object and achieve a defined target.
    \fetchpick{} (\myfig{fig:env_pick}) requires picking up an object from the table and raising it to a target location.
    On the other hand, \fetchpush{} (\myfig{fig:env_push}) requires pushing an object on the table and moving it to a target location.
    We use 10k state-action pairs provided by~\cite{lee2021generalizable} as our expert data for both \fetchpick{} and \fetchpush{}, which contains 303 and 185 trajectories, respectively.
    We set the noise level $p$ to 0.2 to create noisy demonstrations.
    
    \item \textbf{Dexterous Manipulation.}
    We leverage the \handrotate{}~\cite{plappert2018multi}
    environment to represent a challenging Dexterous Manipulation task.
    The task requires controlling a dexterous hand and in-hand rotates a block to a target orientation (\myfig{fig:env_hand}).
    The environment takes 68 dimension states and outputs 20 dimension actions, which is high-dimensional compared to the commonly-used environments in imitation learning.
    We collect 515 trajectories with 10k state-action pairs as our expert data. 
    We set the noise level $p$ to 0.4 to create noisy demonstrations.
    
    \item \textbf{Locomotion.} 
    We leverage the \walker{}~\cite{brockman2016openai}
    environment to represent the locomotion tasks. 
    This environment requires controlling a bipedal agent to travel toward the x-axis direction as fast as possible while maintaining the balance  (\myfig{fig:env_walker}).
    If the agent loses its balance, \eg the height of the agent is too low, the episode would terminate before the maximum number of steps is reached.
    We use 20 trajectories with 20k state-action pairs provided by~\cite{pytorchrl} as our expert data.
    We set the noise level $p$ to 0.2 to create noisy demonstrations.
\end{itemize}

\subsection{Baselines}
\label{sec:exp_baseline}

To evaluate the effectiveness of the proposed method, we compare our \method{} with other offline imitation learning baselines for noisy demonstrations.
Under this problem formulation, the baselines should not require environmental interactions or additional annotations.
Therefore, online methods such as~\cite{wu2019imitation, Tangkaratt2020RobustIL, zhang2021confidence, brown2019extrapolating, brown2020better} are not applicable and we compare our method with the following baselines:

\begin{itemize}
    \item \textbf{BC.} 
    Behavioral Cloning (BC) is a straightforward approach to offline imitation learning. It learns a policy that directly maps from states to actions by imitating the behavior demonstrated in training data using supervised learning. However, BC struggles to deal with noisy demonstrations since the policy tends to overfit noisy data, making it challenging to learn the underlying dynamics accurately.
    
    \item \textbf{Ensemble BC.} 
    To mitigate the impact of noisy demonstrations, one effective strategy is to employ ensemble techniques. 
    Ensemble BC extends the basic BC approach by training several policies and aggregating their outputs.
    Ensemble BC reduces the bias and variance of the prediction and thus improves the overall resilience to noise perturbation.
    Implementation details for Ensemble BC, and a further study for ensemble policies are elaborated in~\mysecref{sec:exp_ensemble} of our supplementary material. 
    
    \item \textbf{BCND.}
    Behavioral Cloning from Noisy Demonstrations (BCND)~\cite{sasaki2020behavioral} aims to learn robust policies from noisy demonstrations containing optimal and sub-optimal behaviors.
    They apply ensemble policies and further design an algorithm to assign weights for each training sample. 
    With the derived weights, the policies are encouraged to capture the behaviors of the major distribution of training samples, which are assumed to be clean and optimal.
\end{itemize}

\subsection{Quantitative Results}
\label{sec:exp_results}

\input{table/main_table}
\input{fig/visualization}
We compare our \method{} with baselines and show the experimental results in~\mytable{table:main}.
The results are separated into two parts: methods with a single policy and methods that leverage the ensemble technique with multiple policies.
We also report the result of DMDR when the ensemble technique is applied (Ensemble \method{}).
We evaluate the agents with 100 episodes and five random seeds on all tasks.
We report the average and standard deviation of the success rate for \fetchpick{}, \fetchpush{}, and \handrotate{} and return for \walker{}.

We observe that the proposed \method{} outperforms baselines whether the ensemble policies are applied or not.
The above results verify the effectiveness of our demonstration filtering and restoration process.
BC struggles with learning from noisy demonstrations, which suggests that the noisy data points severely hinder performance.
By aggregating multiple BC policies, Ensemble BC improves its performance compared to BC, and the standard deviations are lower than those from BC in all tasks, which infers that ensemble policies are more robust to noisy demonstrations.
For BCND, it slightly outperforms Ensemble BC, especially in \handrotate{}. However, it produces a large standard deviation in most of the tasks since the weight-assigning mechanism is affected by the noisy samples at the early training stage.
The learning curves for these experiments are presented in~\mysecref{sec:progress} of the supplementary material.

\subsection{Visualization Results }
To demonstrate the effectiveness of our restoration approach intuitively, we conduct a visualization experiment using t-SNE projections with noisy demonstrations from the \fetchpick{} environment.
As illustrated in~\myfig{fig:visualization}, clean, noisy, and restored samples are represented in green, red, and blue, respectively.
The locations of the clean samples are fixed for easier comparison.
From~\myfig{fig:n_state} and~\myfig{fig:r_state}, 
it can be observed that the restored states (blue) are distributed significantly closer to the clean states (green) than the noisy states (red) prior to the restoration process.
Similar results can be observed for the actions by comparing~\myfig{fig:n_action} and~\myfig{fig:r_action}.
These results confirm that our restoration process effectively recovers distorted states and actions. Consequently, the policy trained on the restored dataset achieves improved performance.

\subsection{Ablation Study for Demonstration Filtering}
\label{sec:exp_filter}
To evaluate the effectiveness of our demonstration filtering design, we ablate the filtering part while fixing the restoration algorithm and the setting for the policy learning.
We compare the filtering approaches listed in the following:

\begin{itemize}
    \item \textbf{Random Filtering.}
    A naive way to do filtering is to randomly label half the states and actions as clean independently. This baseline serves as a bottom line for filtering methods.

    \item \textbf{Autoencoder (AE).}
    Two autoencoders are learned for states and actions, respectively.
    Since autoencoders tend to capture the major behaviors of training data, samples with higher reconstruction losses are considered outliers and labeled as noisy.

    \item \textbf{Local Outlier Factor (LOF).}
    LOF estimates the local deviation of a data point with respect to its neighbors given a dataset.
    Specifically, we filter samples by computing the LOF score based on $k$-nearest neighbors and then identifying a sample as an outlier if the score is large.
    
    \item \textbf{Ours.}
    Our method calculates the LOF scores for each data point based on the features derived from the autoencoders.
    This method considers global representations with the autoencoders and local deviations of samples with the LOF algorithm.
\end{itemize}

The results of filtering ablation are shown in~\mytable{table:filter}.
We note that employing the naive random filtering method demonstrates an improvement in success rate alongside a reduction in variance.
We hypothesize that the improvement results from the restoration process.
Even though the training data for diffusion models and noise predictors are still noisy because of random filtering, the diffusion models can still improve the quality of samples by restoration.
The above results strengthen the motivation of our filter-and-restore framework.

Utilizing Autoencoder (AE) or Local Outlier Factor (LOF) individually for filtering demonstrations shows a respectable improvement.
However, the autoencoders only capture the majority of all training samples and can only better describe the global behaviors of the demonstrations.
On the other hand, the Local Outlier Factor (LOF) only considers local features and is unaware of global behaviors. 
These restrictions hold back their ability to filter out noisy samples accurately.

In contrast, our proposed approach, which combines AE and LOF, capitalizes on the strengths of both methods. 
By integrating global and local feature representations, our method surpasses the performance of its components, leading to superior results.
\input{table/ablation_filter}

\subsection{Ablation Study for Trusted Data Fraction}
\label{sec:TrustedFraction}
In the demonstration filtering stage, we rank all samples by their Local Outlier Factor (LOF) scores and keep the lowest‐ranked fraction as trusted data for training the diffusion model. This follows the \emph{majority-clean assumption} widely used in unsupervised anomaly detection, which states that at least half of the dataset is clean.
Accordingly, our default choice is to retain the lowest 50\,\% of the samples.  
To confirm that this design remains effective in practice, we vary the trusted data fraction~$\tau$ from $30\%$ to $90\%$ and measure the downstream policy performance on \fetchpick{}, keeping all other components of \method{} unchanged. 

\input{table/ablation_trustedData}

The results in~\mytable{table:trusted_fraction} show a steady improvement in performance as the trusted data fraction decreases from $90\%$ to $30\%$, but performance drops sharply when fewer than $50\%$ of the samples are retained. Higher values of $\tau$ preserve more data and increase diversity, but also include noisier samples that negatively affect the quality of the learned restorations. On the other hand, lower values of $\tau$ remove too many samples, reducing coverage and limiting the downstream policy to learn. The performance peak is exactly at $\tau = 50\%$. This is the largest subset where the clean and noisy separation remains trustworthy, matching the majority-clean assumption that at least half of the dataset is uncontaminated. We keep $\tau = 50\%$ as a robust and task-agnostic setting for all experiments in \method{}.

\subsection{Ablation Study for Demonstration Restoring}
\label{sec:exp_restore}
To verify the effect of demonstration restoration and evaluate the designs of our restoration method proposed in~\mysecref{sec:demo_restore}, we compare different strategies that deal with samples from the noisy subsets.

We compare our method with the following baselines and variants on \fetchpick{} and \walker{} environments:
\begin{itemize}
    \item \textbf{Random forest regressor}:
    Random forest regressor makes predictions based on the predictions from multiple decision trees.
    We train the regressor using the clean state-action pairs from $D_{(\hat{s}, \hat{a})}$ and use the trained regressor to predict the noisy states or actions given the corresponding clean actions or states.
    \item \textbf{Generation}:
    Given the diffusion models learned in the filtering stage, this baseline directly generates states/actions based on the corresponding actions/states. We compare our restoration method with this baseline to verify the benefits of restoring noisy samples instead of generating samples from isotropic Gaussian noises directly.
    \item \textbf{Ours w/o predictor}:
    To verify the effectiveness of our noise timestep predictors, we apply restoration with a fixed noise timestep $t$ for the diffusion model. Given that the total number of diffusion steps $T$ is $100$, we set the fixed noise timesteps as $50$.
    \item \textbf{Ours w/o $t_{\text{thres}}$}:
    As described in~\mysecref{sec:method_predictor}, the predicted noise timestep $t^{*}$ from predictors can be used to filter noisy demonstrations by thresholding.
    To evaluate the above design, we employ a variant of our method that does not apply thresholding and directly restores the noisy samples based on $t^{*}$.
    \item \textbf{Ours}:
    Our restoration method utilizes the predicted noise predictors to predict the noise timestep for each noisy sample and further sets a threshold $t_{\text{thres}}$ to filter samples with smaller noise timesteps before restoring them.
\end{itemize}
\input{table/ablation_restore}

The results of restoration ablation are shown in~\mytable{table:restore}.
The random forest regressor is outperformed by other diffusion-model-based restoration methods, which verifies the effectiveness of using diffusion models for recovering data.
We observe that all diffusion-model-based methods perform similarly on \fetchpick{}, including the augmentation with diffusion models baseline. The results infer that a well-trained diffusion model can directly generate informative training data from noises in this environment. Therefore, all of the diffusion-model-based methods perform well regardless of whether the noise predictors are included.

On the other hand, our restoration method outperforms all baselines and variants in \walker{}.
The results indicate that generating data from noise and restoring data without noise predictors can not restore the noisy sample effectively and highlight the importance of utilizing noise predictors and the thresholding method.

\subsection{Imitation Learning Algorithms with \method{}}
\label{sec:exp_il_alg}

Using \method{} to restore demonstrations not only benefits policy learning of BC but also other imitation learning algorithms.
Here, we evaluate three offline imitation learning algorithms on the \fetchpick{} environment to compare the performance when using noisy demonstrations and the demonstrations restored by our \method{}.

\begin{itemize}
    \item \textbf{BC.} Behavioral cloning (BC) is a straightforward baseline that learns a policy to map states to actions directly using the mean square error (MSE) for training.
    \item \textbf{Implicit BC} 
    \cite{florence2022implicit} utilizes an energy-based model (EBM) to train an implicit behavior-cloning policy, which models the expert policy. 
    The training of the energy-based model employs the InfoNCE loss, as described in \cite{oord2018representation}.
    
    \item \textbf{Diffusion Policy} 
    Diffusion Policy~\cite{pearce2022imitating, chi2023diffusionpolicy} learn a conditional diffusion model using diffusion loss to predict actions given the observed states.
    During inference, the diffusion model takes the current state as a condition and gradually denoises the action from noise with the reverse diffusion process.
\end{itemize}

As shown in~\mytable{table:alg}, all algorithms struggle to learn directly from the noisy demonstrations but can significantly improve using the restored demonstrations from our \method{}, while Implicit BC and Diffusion Policy are even more sensitive to noisy demonstrations than BC. 
\method{} benefits these imitation learning algorithms by restoring the noisy demonstrations and results in superior performances with more stable training progressing (lower standard deviation), enabling broader applications of real-world scenarios.
\input{table/ablation_algorithm}


\subsection{Ablation Study on Noise Types}
\label{sec:noise_type}

To verify that the proposed \method{} framework is not restricted to a single noise model, we repeat the \fetchpick{} experiment under various noise types that together cover light-tailed, heavy-tailed, biased, and mixture-sourced noise. All settings keep the overall noise level $p=0.2$ and fix the noise standard deviation  $\sigma_s = \sigma_a = \tfrac{1}{6}$. The results are shown in~\mytable{table:noiseType_ablation}.

\begin{itemize}
\item \textbf{Unbiased noise.} 
\method{} maintains high success under Gaussian and Laplacian noise, indicating that the filter-and-restore pipeline is not tied to light-tailed assumptions. 
However, under uniform noise \method{} (72.0\%) is \emph{comparable} to BC (74.8\%).
LOF relies on spotting points whose local density is much lower than that of their $k$ neighbours~\cite{breunig2000lof}.
Uniform noise fills the space evenly, so every sample has neighbors at a similar density, which hinders the performance of density-based anomaly detection~\cite{campos2016evaluation}.

\item \textbf{Biased noise.} 
To simulate systematic sensor drift, we inject a constant offset of  $+0.4$ into every corrupted sample, producing \emph{Gaussian (biased)} and \emph{Uniform (biased)} settings.  

BC struggles with these shifted demonstrations.
Our goal is to test whether DMDR can detect displacement and restore demonstrations to the clean manifold.  
Significant performance improvements confirm \method{}’s ability to correct systematic shifts that conventional imitation learning suffers.

\item \textbf{Multi-source noise.} 
Real sensor errors rarely come from a single mechanism, so we also evaluate two mixture settings:  
\textit{Gaussian+Uniform} (half of the corruptions sampled from a zero-mean Gaussian, half from a zero-mean Uniform range) and  
\textit{Gaussian+Gaussian} (a bimodal Gaussian with peaks at $\pm0.4$).  

\method{} improves BC's performance \emph{without knowing to underlying noise distribution}.  The result confirms that \method{} can handle realistic, multi-sourced noise.
\end{itemize}

These results confirm that \textbf{\method{} generalizes beyond Gaussian noise} and is resilient to a spectrum of realistic noise, with the only limitation arising in the edge case of uniform noise—where density-based anomaly detection performs poorly. This suggests future work to explore outlier detection strategies leveraging temporal or model-based structure.

\input{table/ablation_noiseType}

\subsection{Effect of Noise Level $p$}
\label{sec:noise_level}

To verify that \method{} is robust to varying noise levels, we evaluate its performance under two noise levels, $p = 0.2$ and $p = 0.4$. The results are shown in~\myfig{fig:noiselevel}, where each task reports the performance of BC and \method{} under both noise levels.

In general, \method{} consistently outperforms BC in all tasks and both noise levels. For instance, in \fetchpick{}, DMDR achieves a significantly higher success rate than BC at both $p = 0.2$ and $p = 0.4$. A similar trend is observed in \fetchpush{} and \walker{}, where increasing $p$ leads to performance degradation for both methods, but DMDR maintains a clear advantage.

The \handrotate{} task, which operates in a high-dimensional state and action space, presents a more challenging scenario. At p = 0.2, the performance gap between DMDR and BC is smaller, as the corruption may not be enough to show a clear difference. However, at p = 0.4, DMDR achieves better performance than BC, suggesting that it remains effective even under more challenging noise conditions.

These results demonstrate that DMDR is robust not only to different types of noise but also to different levels of noise. This robustness is crucial in real-world applications, where the actual corruption rate is unknown and may vary across tasks and environments.

\input{fig/noise_level}

%% file: fig/env.tex
\begin{figure*}[!t]
    \centering
    \subfloat[][\fetchpick{}]
    {\includegraphics[width=0.22\textwidth]{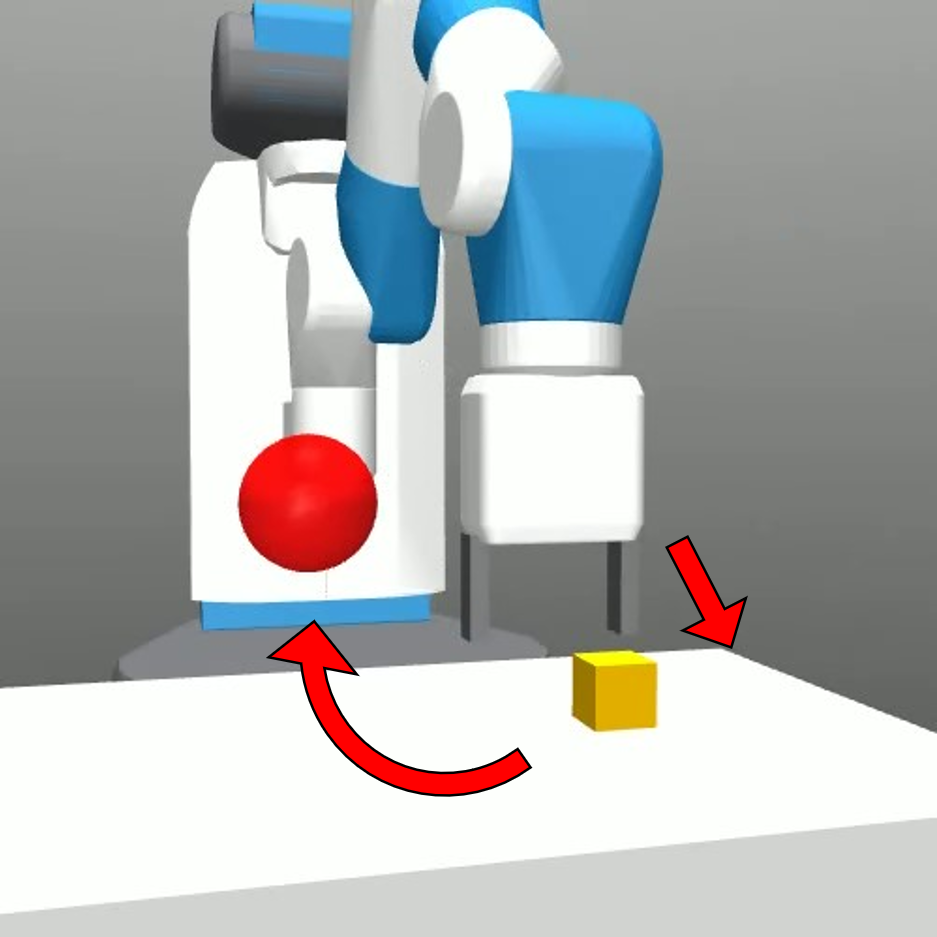}
    \label{fig:env_pick}}
    \hspace{0.2cm}
    \subfloat[][\fetchpush{}]
    {\includegraphics[width=0.219\textwidth]{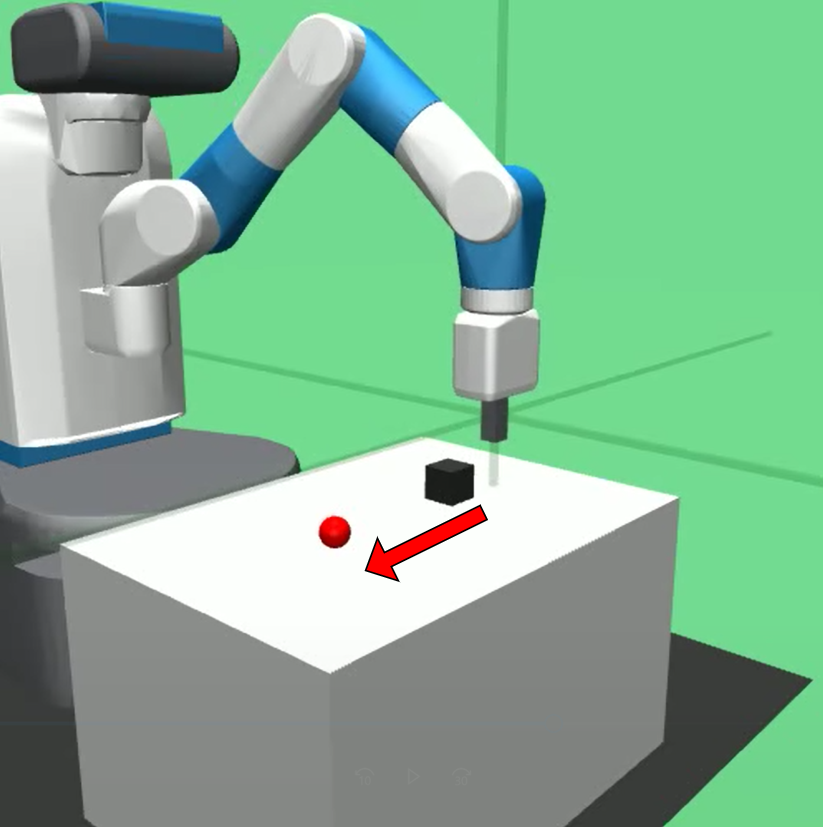}
    \label{fig:env_push}}
    \hspace{0.2cm}
    \subfloat[][\handrotate{}]
    {\includegraphics[width=0.22\textwidth]{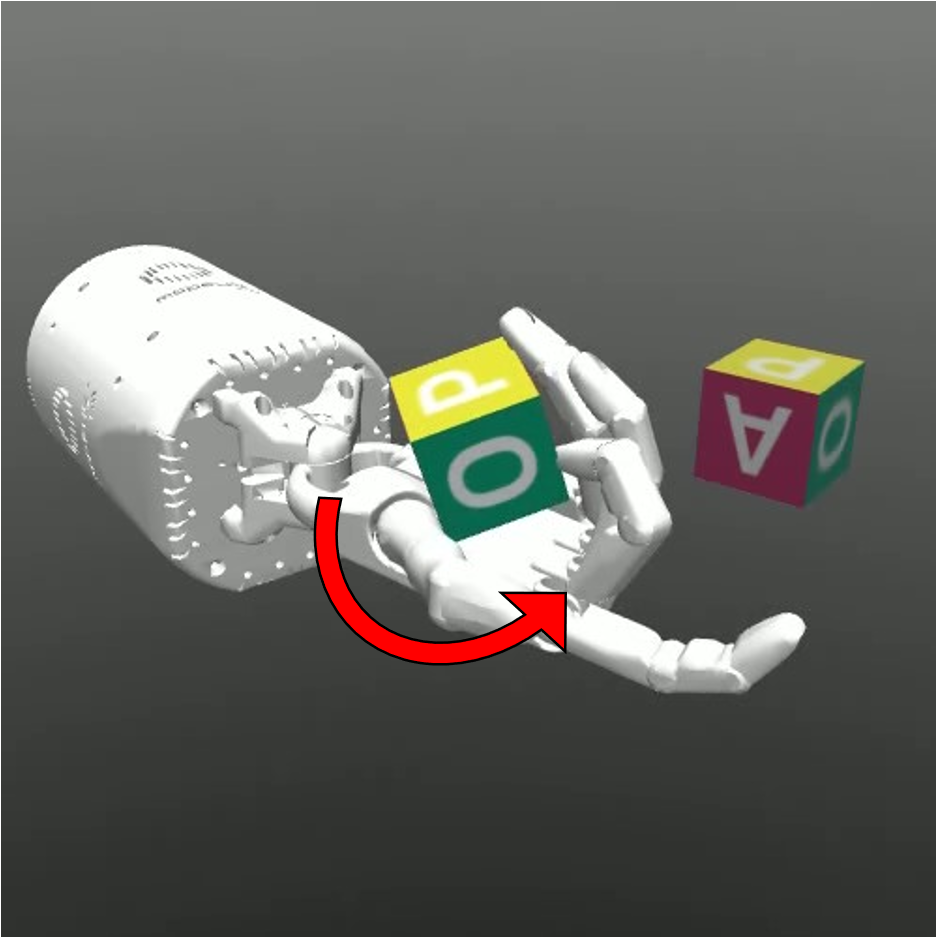}
    \label{fig:env_hand}}
    \hspace{0.2cm}
    \subfloat[][\walker{}]
    {\includegraphics[width=0.22\textwidth]{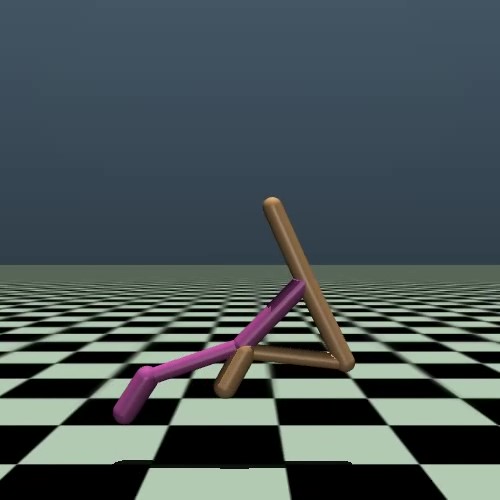}
    \label{fig:env_walker}}
    \caption[]{ 
    \textbf{Environments \& Tasks.}
    \textbf{(a)-(b) \fetchpick{} and \fetchpush{}}:
    The robot arm manipulation tasks employ a 7-DoF Fetch robotics arm to pick up/push an object from the table and move it to a target location.
    \textbf{(c) \handrotate{}}:
    This dexterous manipulation task requires a Shadow Dexterous Hand to in-hand rotate a block to a target orientation.
    \textbf{(d) \walker{}}:
    These locomotion tasks require 
    learning agents
    to walk as fast as possible while maintaining their balance.}
    \label{fig:environment}
\end{figure*}

%% file: table/main_table.tex
\begin{table*}
\centering
\caption[]{Overall experiment results}
\scalebox{1.2}{\begin{tabular}{@{}cccccc@{}}
\toprule
\textbf{Method} &
\# of policies &
\fetchpick &
\fetchpush &
\handrotate &
\walker \\
\cmidrule{1-6}
BC &
1 &
44.38\% $\pm$ 11.02\% &
67.97\% $\pm$ 4.81\% &
45.56\% $\pm$ 6.15\% &
4456.8 $\pm$ 1051.1 \\
\method{} (Ours)&
1 &
\textbf{90.52}\% $\pm$ 4.83\% &
\textbf{79.64}\% $\pm$ 5.30\% &
\textbf{51.70}\% $\pm$ 5.85\% &
\textbf{5066.4} $\pm$ 886.4 \\
\cmidrule{1-6}
Ensemble BC &
5 &
51.36\% $\pm$   6.67\% &
72.25\% $\pm$ 3.20\% &
51.03\% $\pm$ 4.33\% &
5170.6 $\pm$ 395.2 \\ 
BCND~\cite{sasaki2020behavioral} &
5 &
52.87\% $\pm$ 12.71\% &
71.01\% $\pm$ 17.98\% &
\textbf{54.97}\% $\pm$ 4.49\% &
5144.8 $\pm$ 739.6 \\ 
Ensemble \method{} (Ours) &
5 &
\textbf{91.80}\% $\pm$ 2.54\% &
\textbf{82.03}\% $\pm$ 2.87\% &
\textbf{55.36}\% $\pm$ 4.43\% &
\textbf{6168.1} $\pm$ 284.9 \\
\bottomrule
\end{tabular}}
\label{table:main}
\end{table*}

%% file: fig/visualization.tex
\begin{figure*}[!t]
    \centering
    \subfloat[][Noisy State]
    {\includegraphics[width=0.21\textwidth]{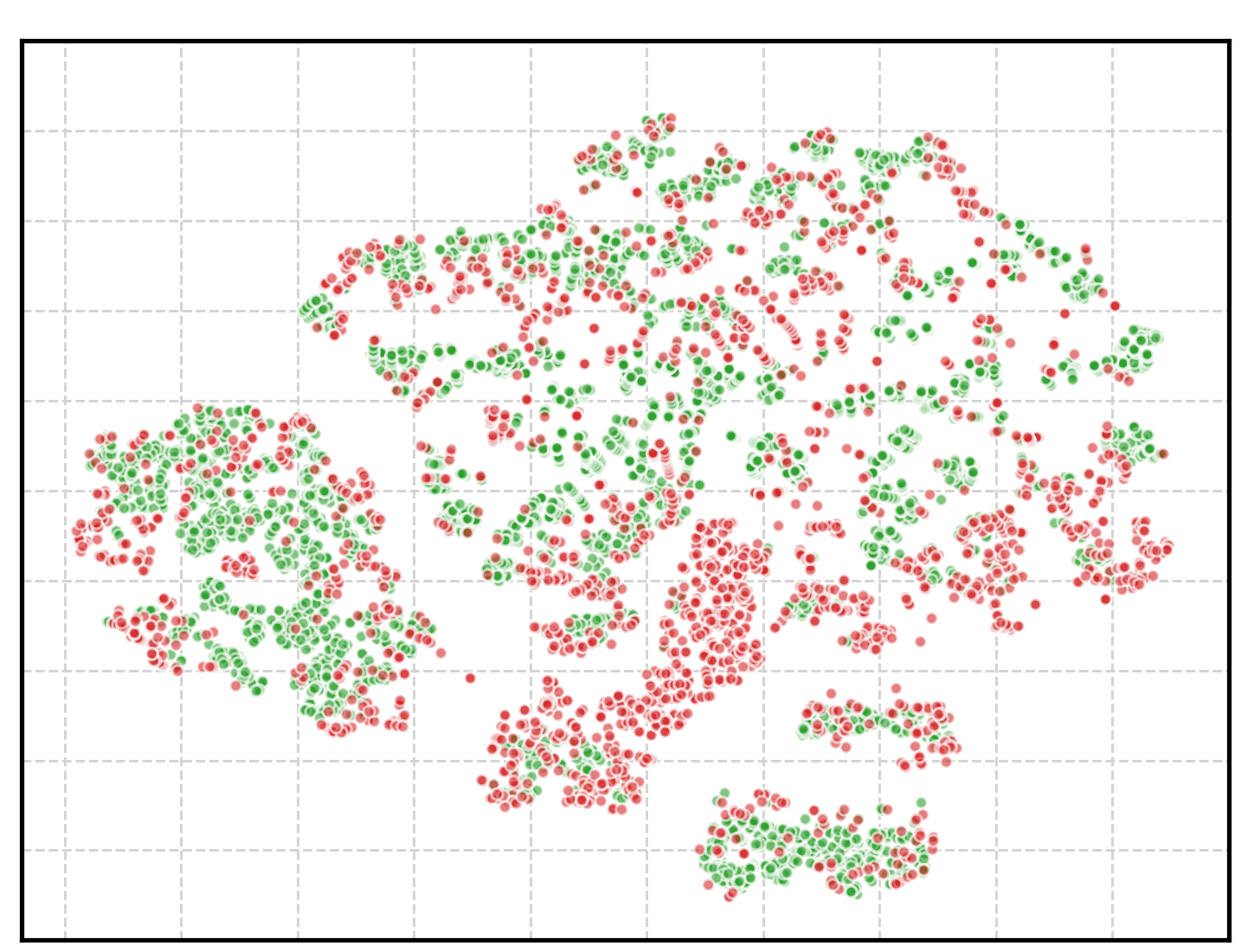}
    \label{fig:n_state}}
    \subfloat[][Restored State]
    {\includegraphics[width=0.21\textwidth]{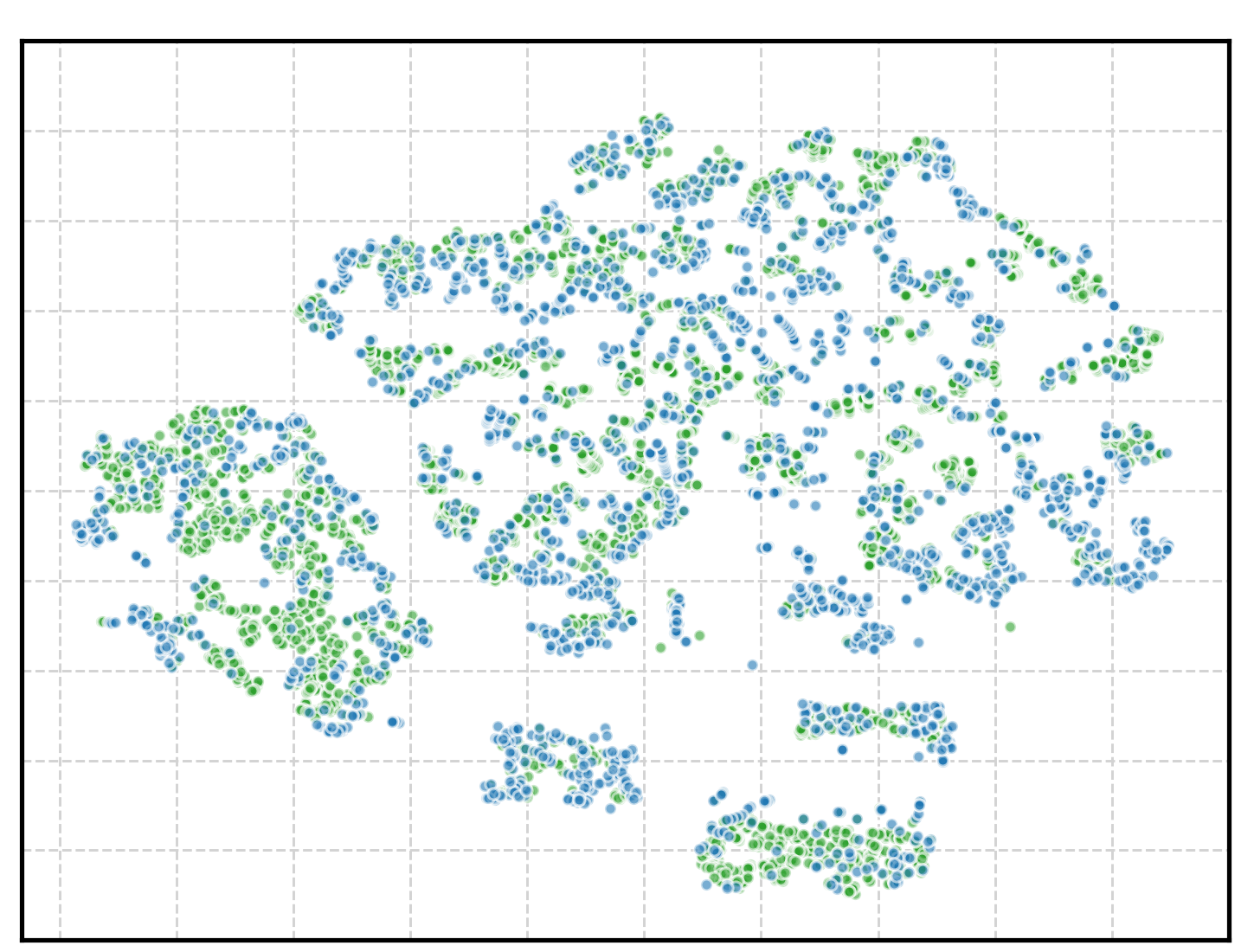}
    \label{fig:r_state}} 
    \raisebox{0.3cm}{\includegraphics[width=0.1\textwidth]{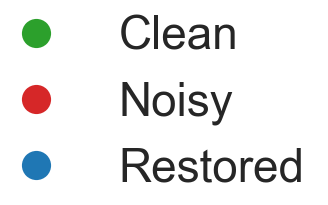}}
    \subfloat[][Noisy Action]
    {\includegraphics[width=0.21\textwidth]{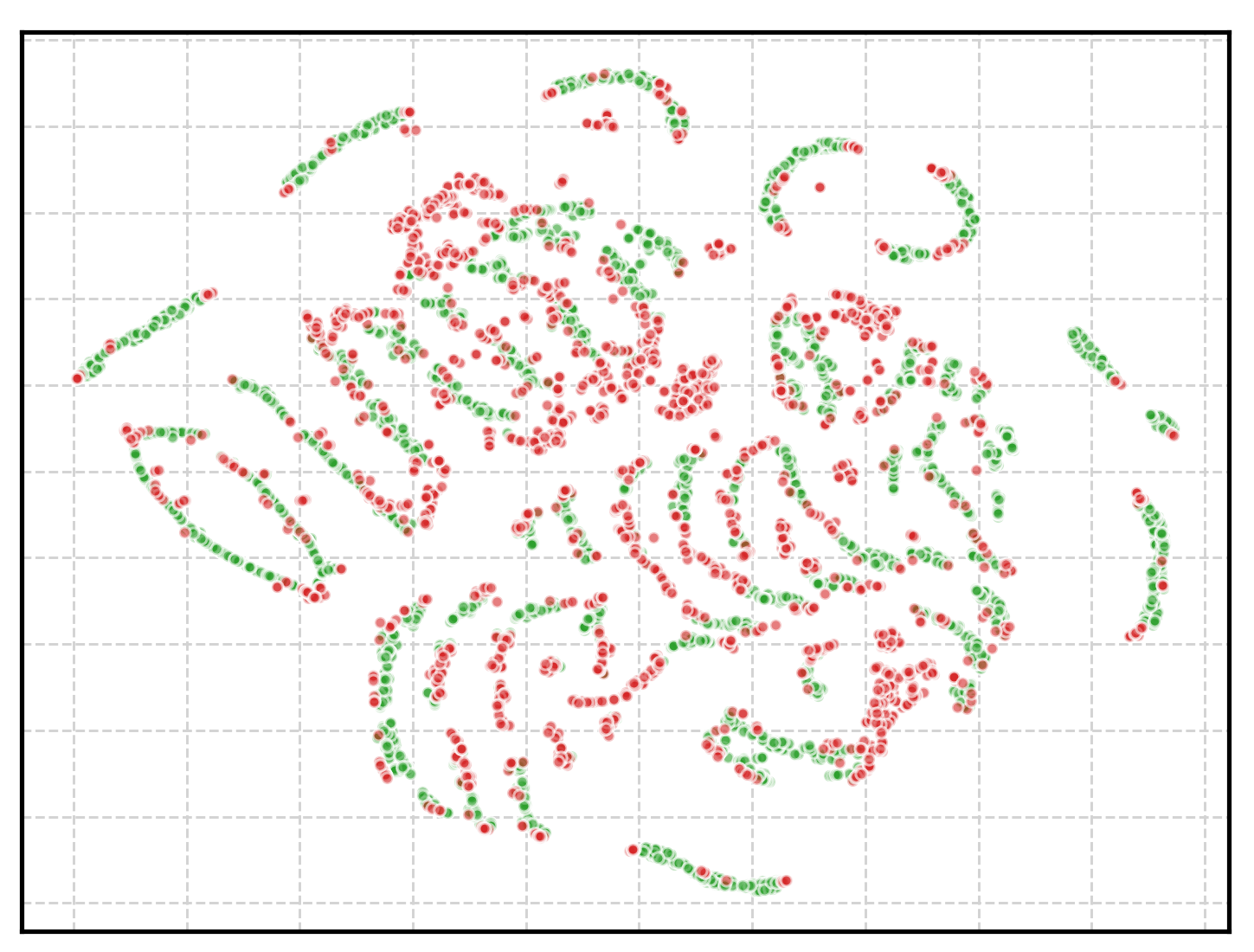}
    \label{fig:n_action}}
    \subfloat[][Restored Action]
    {\includegraphics[width=0.21\textwidth]{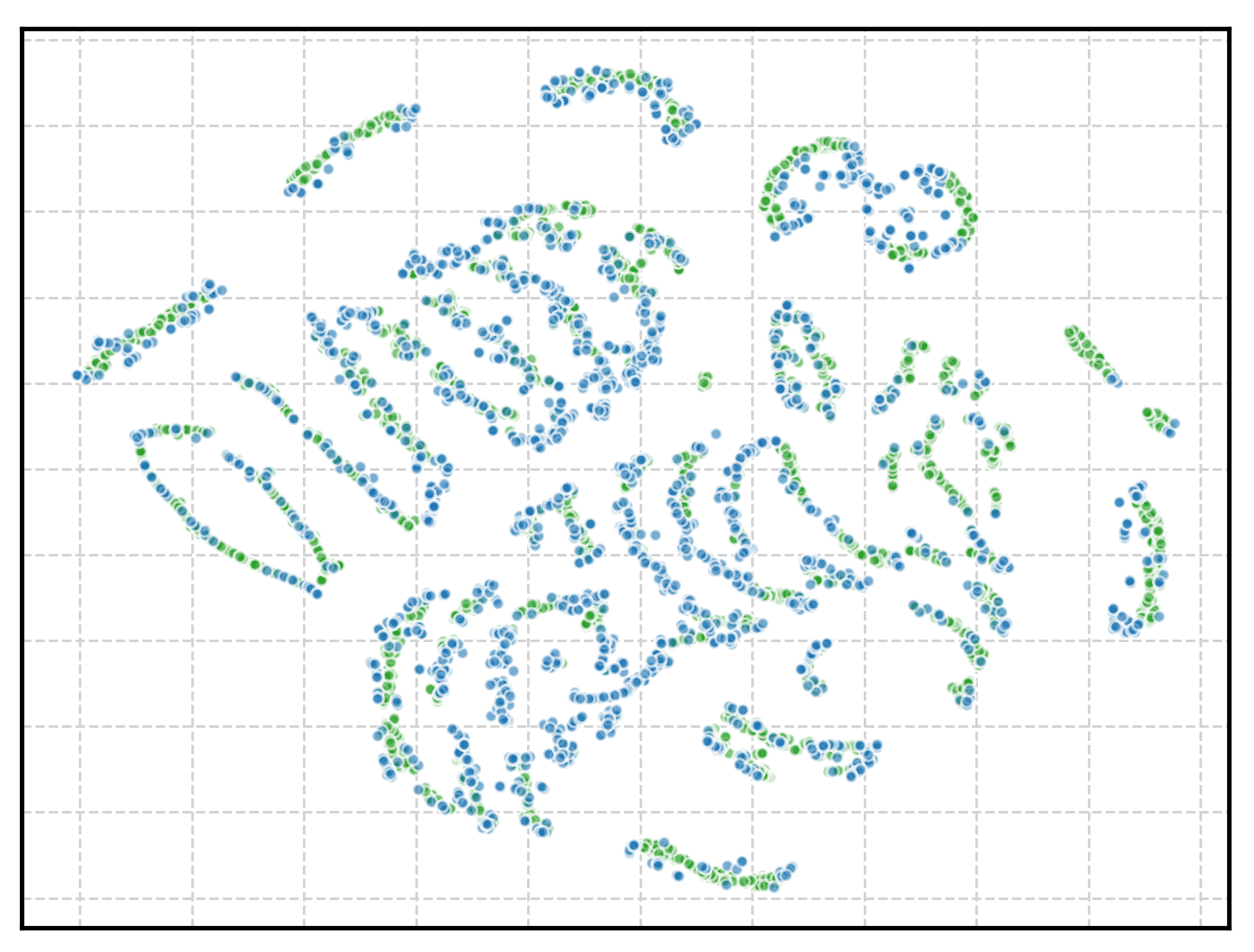}
    \label{fig:r_action}}
    \caption[]{
    \textbf{Visualization Results on \fetchpick{}.} We use t-SNE to project noisy demonstrations from the \fetchpick{} environment. 
    The clean, noisy, and restored samples are represented in green, red, and blue, respectively.
    From~\myfig{fig:n_state} and~\myfig{fig:r_state}, it can be observed that the restored states (blue) are distributed significantly closer to the clean states (green) than the noisy states (red) prior to the restoration process.
    Similar results can be observed for the actions by comparing~\myfig{fig:n_action} and~\myfig{fig:r_action}.}
    \label{fig:visualization}
\end{figure*}

%% file: table/ablation_filter.tex
\begin{table}[t]
\small
\centering
\caption[]{Demonstration filtering ablation study}
\scalebox{1.1}{\begin{tabular}{@{}ccc@{}}
\toprule
\textbf{Method} & 
\# of Samples &
Success Rate \\
\cmidrule{1-3}
Noisy Data &
10000 &
45.40\% $\pm$ 11.33\% \\
Random Filtering & 
7505 & 
51.20\% $\pm$ 6.80\% \\ 
AE &
7313 & 
86.00\% $\pm$ 3.97\% \\ 
LOF & 
7311 &
76.75\% $\pm$ 27.93\% \\
Ours & 
7362 &
\textbf{91.80\% $\pm$ 4.09\%} \\
\bottomrule
\end{tabular}}
\label{table:filter}
\end{table}

%% file: table/ablation_trustedData.tex
\begin{table}[h]
\small
\centering
\caption{Policy performance on Pick under different trusted data fractions $\tau$.}
\label{table:trusted_fraction}
\scalebox{1.1}{
\begin{tabular}{@{}ccc@{}}
\toprule
$\tau$ & Discarded samples & Success Rate \\
\midrule
90\,\% & 107   & 53.80\% ± 8.35\% \\
80\,\% & 414   & 70.60\% ± 6.03\% \\
70\,\% & 993   & 84.60\% ± 7.23\% \\
60\,\% & 1760 & 87.80\% ± 6.72\% \\
\textbf{50\,\%} & \textbf{2,638} & 
\textbf{90.52\% ± 4.83\%} \\
40\,\% & 3760 & 32.60\% ± 4.88\% \\
30\,\% & 4993 & 3.00\% ± 1.41\% \\
\bottomrule
\end{tabular}}
\end{table}

%% file: table/ablation_restore.tex
\begin{table}[t]
\small
\centering
\caption[]{Demonstration restoration ablation study}
\scalebox{0.98}{\begin{tabular}{@{}cccc@{}}
\toprule
\multirow{2}{*}{\textbf{Method}}
& \fetchpick{} & \walker{} \\
& (Success Rate) & (Return) \\
\cmidrule{1-3}
Random forest regressor &
65.00\% $\pm$ 12.71\% &
255.3 $\pm$ 48.1 \\
Generation & 
\textbf{89.89}\% $\pm$ 5.13\% &
3452.6 $\pm$ 1528.8 \\ 
Ours w/o predictor & 
88.20\% $\pm$ 4.82\% & 
2702.9 $\pm$ 1281.3 \\
Ours w/o $t_{\text{thres}}$ & 
89.00\% $\pm$ 4.00\% &
4261.4 $\pm$ 857.2 \\
Ours & 
\textbf{90.52}\% $\pm$ 4.83\% &
\textbf{5066.4} $\pm$ 886.4 \\
\bottomrule
\end{tabular}
}
\label{table:restore}
\end{table}

%% file: table/ablation_algorithm.tex
\begin{table}
\small
\centering
\caption[]{}
Imitation Learning Algorithms with \method{}
\scalebox{1.1}{\begin{tabular}{@{}ccccc@{}}
\toprule
\textbf{Algorithm} & 
Noisy Demo. &
Restored Demo. 
\\
\cmidrule{1-3}

BC &
44.38\% $\pm$ 11.02\% & 
\textbf{90.52}\% $\pm$ 4.83\% \\

Implicit BC & 
3.00\% $\pm$ 3.24\% & 
\textbf{44.56}\% $\pm$ 6.98\%\\ 

Diffusion Policy &
25.80\% $\pm$ 2.59\% & 
\textbf{97.40}\% $\pm$ 2.19\% \\
\bottomrule
\end{tabular}}
\label{table:alg}
\end{table}

%% file: table/ablation_noiseType.tex
\begin{table}[ht]
\small
\centering
\caption{Success rate on \textsc{FetchPick} under different
noise types.}
\label{table:noiseType_ablation}
\resizebox{\columnwidth}{!}{%
\begin{tabular}{lcc}
\toprule
\textbf{Noise type} & \textbf{BC} & \textbf{DMDR} \\
\midrule
\multicolumn{3}{l}{\textbf{Unbiased noise}}\\
Gaussian            & $44.38\%\ \pm\ 11.02\%$ & $\mathbf{90.52\%\ \pm\ 4.83\%}$ \\
Laplacian           & $54.00\%\ \pm\ 7.21\%$  & $\mathbf{87.40\%\ \pm\ 2.70\%}$ \\
Uniform             & $\mathbf{74.80\%\ \pm\ 6.02\%}$ & $\mathbf{72.00\%\ \pm\ 13.51\%}$ \\
\midrule
\multicolumn{3}{l}{\textbf{Biased noise} ($+0.4$ offset)}\\
Gaussian (biased)   & $8.80\%\ \pm\ 4.32\%$   & $\mathbf{76.40\%\ \pm\ 19.76\%}$ \\
Uniform (biased)    & $0.00\%\ \pm\ 0.00\%$   & $\mathbf{35.80\%\ \pm\ 14.24\%}$ \\
\midrule
\multicolumn{3}{l}{\textbf{Mixed-source noise}}\\
Gaussian+Uniform    & $17.80\%\ \pm\ 12.56\%$ & $\mathbf{82.20\%\ \pm\ 7.19\%}$ \\
Gaussian+Gaussian   & $45.40\%\ \pm\ 5.03\%$  & $\mathbf{89.20\%\ \pm\ 7.09\%}$ \\
\bottomrule
\end{tabular}%
}
\end{table}

%% file: fig/noise_level.tex
\begin{figure}[h]
    \centering
    \vspace{-0.3cm}
    \includegraphics[width=1.0\linewidth]{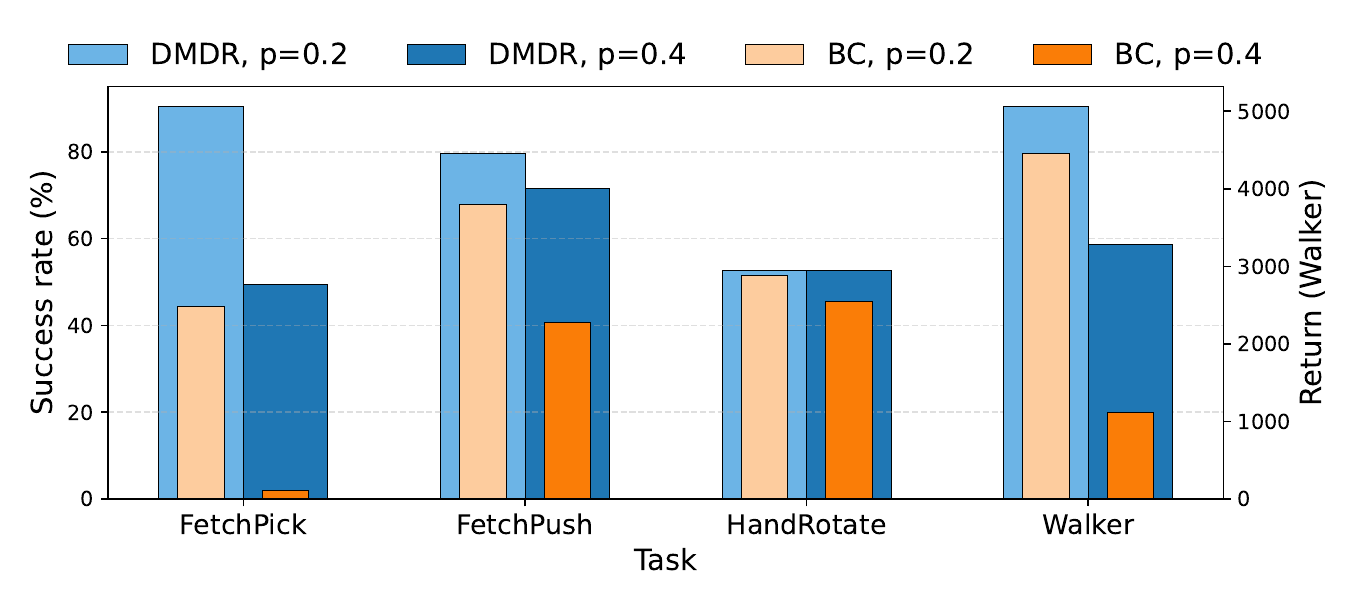}  
    \vspace{-0.8cm}
    \caption{Sensitivity to noise level $p$. Success rate for manipulation tasks; episodic return for \walker.}
    \vspace{-0.5cm}
    \label{fig:noiselevel}
\end{figure}

%% file: text/5_discussion.tex
\section{Discussion and Limitation}
\label{sec:limitation}
In this paper, we propose an offline imitation learning framework (\method{}) that enables imitation learning methods to learn from noisy demonstrations.
Our \method{} first filters clean samples from the demonstrations and then learns diffusion models to restore the noisy samples.
The experiments show that \method{} outperforms baselines for learning from noisy demonstrations in various tasks, including robot arm manipulation, dexterous manipulation, and locomotion.
Ablation studies investigate the effect of the filtering method and restoration method.
We also show that our \method{} can be applied to various imitation learning algorithms to verify the effectiveness of our proposed method. 
Most importantly, \method{} remains robust under complex, mixed-source noise without any prior knowledge of the
underlying noise distribution.

Despite this broad robustness, one limitation is that density-based anomaly detectors, such as LOF, are restricted in their performance on an edge case when the noise distribution is uniform. Leveraging temporal or model-based structures could be a worthwhile future direction.

Another potential limitation of \method{} is that the diffusion models are trained on the clean subset $D_{(\hat{s}, \hat{a})}$.
When this subset is small, the restored trajectories may lack diversity and fail to recover the remaining demonstrations.
Transitions flagged as outliers, \((s',a')\), are currently \emph{discarded}, preventing any potentially useful information from reaching the learner. A promising direction is to recycle these rejected transitions so that every sample, clean or noisy, contributes to learning and improves overall data efficiency.

Lastly, our current pipeline does not attempt to filter or correct demonstrations that are sub-optimal yet uncorrupted; extending DMDR to explicitly handle varying levels of optimality is left for future work.

%% file: text/appendix.tex
\section*{Supplementary Materials}

\section{Algorithm}
Our proposed framework \method{} consists of two stages.
(1) \textbf{Demonstration filtering}: We learn the global features of the expert distribution with a pair of autoencoders $\phi_{s}$ and $\phi_{a}$, and then apply Local Outlier Factor (LOF) to filter clean samples based on local densities. The algorithm for demonstration filtering is detailed in~\myalgo{alg:filter}.
(2) \textbf{Demonstration restoration}:
The conditional diffusion models $\theta_{s}$ and $\theta_{a}$ learn to restore state/action while the noise timestep predictors $\psi_{s}$ and $\psi_{a}$ aim to predict the noise timestep of the noisy state $s'$ and actions $a'$ for restoration.
The algorithm for demonstration restoration is detailed in~\myalgo{alg:restore}.
\label{sec:algorithm}
\input{algo/filter}
\input{algo/restore}

\section{Training Details}
\label{sec:app_details}
We describe the training details of training in this section, 
including the model architectures, hyperparameters, and computation resources.

\subsection{Model Architecture}
\label{sec:model_archi}
We elaborate the model architecture for each component of our \method{} in~\mytable{table:architecture}. For fair comparisons, the policy architectures of baselines used in this paper follow identical designs. 
\input{table/architecture}

\subsection{Hyperparamters}
\label{sec:app_hyperparamters}

We report the hyperparameters of learning the component used for all the methods on all the tasks in~\mytable{table:hyperparameter}.
We use the Adam optimizer~\cite{kingma2014adam} and a batch size of $128$ for all the methods on all the tasks.
Also, we use linear learning rate decay for all policy models.
The number of diffusion steps is $100$ for conditional diffusion models, noise predictors, and Diffusion Policies.

\input{table/hyperparameter}

\subsection{Computation Resource}
\label{sec:app_compute}

We conducted all the experiments on the following three workstations:
\begin{itemize}
    \item M1: ASUS WS880T workstation with an Intel Xeon W-2255 (10C/20T, 19.3MiB, 3.70GHz) 20-core CPU, 125GB memory, two NVIDIA GeForce RTX 3080 Ti GPUs
    \item M2: ASUS WS880T workstation with an Intel Xeon W-2255 (10C/20T, 19.3M, 3.7GHz) 20-core CPU, 64GB memory, two NVIDIA RTX 4070 Ti GPUs      
\end{itemize}

\newpage
\section{Ablation Study for Ensemble Methods}
\label{sec:exp_ensemble}

Ensemble policies learn $n$ policy variants and aggregate the predictions.
To improve the robustness using ensemble techniques, randomness among the policies is essential for learning distinct features.
A widely adopted method is to sample different training data for each policy.
We use BC policies to evaluate three ways to sample the training data that introduce randomness for ensemble techniques.

\begin{itemize}
    \item \textbf{Split.}
    In~\cite{sasaki2020behavioral}, they split the demonstration dataset into $n$ parts. Each policy learns from a distinct and disjoint subset of the demonstration data.
    The method provides diversity in the ensemble with the cost of fewer training data for each policy.
    
    \item \textbf{Sample with replacement.}
    In~\cite{brantley2019disagreement}, they sample data with replacement from the demonstration for each policy.
    The number of training data for each policy is identical, but the content of training data is different due to replacement after sampling.
    
    \item \textbf{Shuffle.}
    All $n$ policies are trained using the entire demonstration dataset, where the randomness is introduced by shuffling the training data. 
    Policies can capture various features by learning from training data with different orders. 
\end{itemize}

In~\mytable{table:ensemble}, we adopt $3$ and $5$ policies for the above ensemble approaches and compare them with a single policy baseline.
In general, the performances improve when the number of policies increases, except for the Split ensemble method.
This could be a result of the lack of training data for each policy since the number of samples decreases as the number of policies increases.
We observe that the Shuffle ensemble method with $5$ policies performs the best on all the tasks.
As the Shuffle ensemble method utilizes every accessible training data for each policy, we can infer that the number of training samples is an important factor for these imitation learning tasks.
Therefore, we use the Shuffle ensemble method with $5$ policies for Ensemble BC, BCND, and Ensemble DMDR in the~\mytable{table:main} of our main paper.
The above results also motivate the employment of demonstration restoration, which aims to correct and utilize as many noisy samples as possible.
\input{table/ablation_ensemble}
\input{fig/training_progress}

\section{Environments Complexities}
In this section, we aim to quantitatively evaluate the complexities of the environments used in this paper.
We test the multimodality of each environment since it is more challenging to learn an effective policy if the expert trajectories contain multimodal actions.
In imitation learning, multimodality can stem from the nature of the task (e.g., different goals executed in arbitrary orders) or variations in expert demonstrations (e.g., achieving the same goal through different paths).
For each environment, we create 10 clusters of states and 10 clusters of actions from expert demonstrations. 
Then, we measure if the states in the same cluster share actions from the same clusters. 
Specifically, we determine the predominant action class for each state cluster and compute the proportion of states belonging to that class.
This ratio, ranging from 0.1 to 1, indicates whether actions within a state cluster are randomly distributed (1/10) or consistently the same (1/1).

The results of each environment are reported below:
\begin{itemize}
    \item \fetchpick{}: 0.6286
    \item \fetchpush{}: 0.5482
    \item \handrotate{}: 0.3035
    \item \walker{}: 0.4240
\end{itemize}
We observe that robot arm manipulation (\fetchpick{} and \fetchpush{}) tasks result in higher majority ratios, which indicates that the expert behaviors are more unimodal.
On the other hand, dexterous manipulation (\handrotate{}) tasks and locomotion (\walker{}) result in lower majority ratios, which means the demonstrations contain more multimodal paths or behaviors for similar goals.
This experiment shows that the environments in this paper exhibit varying degrees of multimodal actions and that the proposed \method{} effectively improves the baseline performance across all environments.

\section{Training Progress}
\label{sec:progress}
We illustrate the training progress of all methods in~\myfig{fig:training_progress}.
We observe that \method{} is more robust and stable during the training, resulting in a lower standard deviation.
In contrast, BC encounters a performance drop when the training progress is affected by the noisy samples, which can be observed on \fetchpick{} and \walker{}.

%% file: algo/filter.tex
\begin{algorithm}[b!]
\caption{Demonstration Filtering} 
\label{alg:filter}
\textbf{Input}: Noisy demonstration $D$ \\
\textbf{Output}: Subsets of demonstration $D_{(\hat{s}, \hat{a})}$, $D_{(s', a')}$, $D_{(\hat{s}, a')}$, and $D_{(s', \hat{a})}$ \\
\textcolor{gray}{// Training}
\begin{algorithmic}[1]
\STATE Randomly initialize a pair of autoencoders $\phi_{s}$ and $\phi_{a}$ 
\FOR{each autoencoder training iteration}
    \STATE Sample $(s, a)$ from $D$
    \STATE Update $\phi_{s}$ using $L^{s}_{\text{rec}}$ from~\myeq{eq:rec_loss_s}
    \STATE Update $\phi_{a}$ using $L^{a}_{\text{rec}}$ from~\myeq{eq:rec_loss_a}
\ENDFOR
\end{algorithmic}

\textcolor{gray}{// Inference}
\begin{algorithmic}[1]
\FOR{each state-action pair $(s, a)$ in $D$}
    \STATE Calculate the state feature $z_s$ with $\phi_{s}$
    \STATE Calculate the action feature $z_a$ with $\phi_{a}$
\ENDFOR
\STATE Apply LOF on \{$z_s$\} and \{$z_a$\} separately
\STATE Initialize empty sets $D_{(\hat{s}, \hat{a})}$, $D_{(s', a')}$, $D_{(\hat{s}, a')}$, and $D_{(s', \hat{a})}$ 
\FOR{each state-action pair $(s, a)$ in $D$}
    \STATE Label state $s$ based on LOF score from \{$z_s$\}
    \STATE Label action $a$ based on LOF score from \{$z_a$\}
    \STATE Append $(s, a)$ to the subset with state/action labels 
\ENDFOR

\STATE \textbf{return} $D_{(\hat{s}, \hat{a})}$, $D_{(s', a')}$, $D_{(\hat{s}, a')}$, and $D_{(s', \hat{a})}$ 
\end{algorithmic}

\end{algorithm}

%% file: algo/restore.tex
\begin{algorithm}[b!]
\caption{Demonstration Restoration} 
\label{alg:restore}

\textbf{Input}: Demonstration subsets $D_{(\hat{s}, \hat{a})}$, $D_{(s', \hat{a})}$ and $D_{(\hat{s}, a')}$ \\
\textbf{Output}: $D_{(\hat{s}, \hat{a})}$ with restored state-action pairs \\
\textcolor{gray}{// Training}
\begin{algorithmic}[1]
\STATE Randomly initialize diffusion models $\theta_{s}$ and $\theta_{a}$ 
\FOR{each diffusion model training iteration}
    \STATE Sample $(\hat{s}, \hat{a})$ from $D_{(\hat{s}, \hat{a})}$
    \STATE Update $\theta_{s}$ using $L^{s}_{\text{diff}}$ from~\myeq{eq:diff_state_loss}
    \STATE Update $\theta_{a}$ using $L^{a}_{\text{diff}}$ from~\myeq{eq:diff_action_loss}
\ENDFOR
\FOR{each noise predictor training iteration}
    \STATE Sample $(\hat{s}, \hat{a})$ from $D_{(\hat{s}, \hat{a})}$
    \STATE Update $\psi_{s}$ using $L^{s}_{\text{pred}}$ from~\myeq{eq:pred_state_loss}
    \STATE Update $\psi_{a}$ using $L^{a}_{\text{pred}}$ from~\myeq{eq:pred_action_loss}
\ENDFOR
\end{algorithmic}

\textcolor{gray}{// Inference}
\begin{algorithmic}[1]
\FOR{each state-action pair $(s', \hat{a})$ in $D_{(s', \hat{a})}$}
    \STATE Calculate the predicted noise timestep $t'$ with $\psi_{s}$
    \IF{$t' < t_{\text{thres}}$}
        \STATE Append $(s', \hat{a})$ to $D_{(\hat{s}, \hat{a})}$ 
    \ELSE
        \STATE Restore the noisy state with $\theta_{s}$ and output $s^{*}$
        \STATE Append $(s^{*}, \hat{a})$ to $D_{(\hat{s}, \hat{a})}$ 
    \ENDIF
\ENDFOR
\FOR{each state-action pair $(\hat{s}, a')$ in $D_{(\hat{s}, a')}$}
    \STATE Calculate the predicted noise timestep $t'$ with $\psi_{a}$
    \IF{$t' < t_{\text{thres}}$}
        \STATE Append $(\hat{s}, a')$ to $D_{(\hat{s}, \hat{a})}$ 
    \ELSE
        \STATE Restore the noisy state with $\theta_{a}$ and output $a^{*}$
        \STATE Append $(\hat{s}, a^{*})$ to $D_{(\hat{s}, \hat{a})}$ 
    \ENDIF
\ENDFOR

\STATE \textbf{return} $D_{(\hat{s}, \hat{a})}$
\end{algorithmic}

\end{algorithm}

%% file: table/architecture.tex
\begin{table*}
\small
\centering
\caption[Model Architectures]{Model Architectures}
\scalebox{1.0}{\begin{tabular}{@{}cccccccc@{}}\toprule
\textbf{Model} & 
\textbf{Component} &
\fetchpick{} &
\fetchpush{} &
\handrotate{} &
\walker{} 
\\
\cmidrule{1-6}
\multirow{3}{*}{State/Action Autoencoder}
& Input Dim. & 16/4 & 16/4 & 68/20 & 17/6 \\
& Hidden Dim. & \multicolumn{4}{c}{[512, 128, 32, 128, 512] for all environments} \\
& Output Dim. & 16/4 & 16/4 & 68/20 & 17/6 \\
\cmidrule{1-6}
\multirow{4}{*}{State/Action Diffusion Model} &
\# Layers & 5 & 5 & 5 & 5 \\
& Input Dim. & 20 & 20 & 88 & 23\\
& Hidden Dim. & 1024 & 1024 & 1024 & 1024 \\
& Output Dim. & 16/4 & 16/4 & 68/20 & 17/6 \\
\cmidrule{1-6}
\multirow{4}{*}{Noise Predictor} &
\# Layers & 5 & 5 & 5 & 5 \\
& Input Dim. & 20 & 20 & 88 & 23\\
& Hidden Dim. & 1024 & 1024 & 1024 & 1024 \\
& Output Dim. & 1 & 1 & 1 & 1 \\
\cmidrule{1-6}
\multirow{4}{*}{Policy} &
\# Layers & 4 & 4 & 4 & 4\\
& Input Dim.  & 16 & 16 & 68 & 17 \\
& Hidden Dim. & 512 & 512 & 512 & 512\\
& Output Dim. & 4 & 4 & 20 & 6 \\
\bottomrule
\end{tabular}}
\label{table:architecture}
\end{table*}

%% file: table/hyperparameter.tex
\begin{table*}
\centering
\small
\caption[Hyperparameters]{Hyperparameters} 
\scalebox{1.0}{\begin{tabular}{@{}cccccc@{}}\toprule
\textbf{Method} & \textbf{Hyperparameter} & 
\fetchpick{} &
\fetchpush{} &
\handrotate{} &
\walker{} \\
\cmidrule{1-6}
\multirow{3}{*}{BC} 
& Batch Size & 128 & 128 & 128 & 128  \\
& Learning Rate & 2e-6 & 5e-7 & 5e-5 & 2e-5 \\
& \# Epochs & 2000 & 2000 & 2000 & 2000 \\
\cmidrule{1-6}
\multirow{3}{*}{BCND~\cite{sasaki2020behavioral}}
& Batch Size & 128 & 128 & 128 & 128 \\
& Learning Rate & 1e-6 & 2e-6 & 5e-5 & 5e-6 \\
& \# Epochs & 2000 & 2000 & 2000 & 2000 \\
\cmidrule{1-6}
\multirow{11}{*}{\method{} (Ours)}
& \textbf{Demonstration Filtering} & & & & \\
& Autoencoder Learning Rate & 1e-6 & 1e-6 & 1e-6 & 1e-6 \\
& Autoencoder \# Epochs & 500 & 2000 & 2000 & 500  \\
& LOF \# Neighbors & 50& 50& 10& 50 \\
\cmidrule{2-6}
& \textbf{Demonstration Restoration} & & & & \\
& Diffusion Model Learning Rate & 1e-4 & 1e-4 & 1e-4 & 1e-4 \\
& Diffusion Model \# Epochs & 8000 & 8000 & 8000 & 8000  \\
& Noise Predictor Learning Rate & 1e-6 & 1e-6 & 1e-6 & 1e-6 \\
& Noise Predictor \# Epochs & 2000 & 2000 & 2000 & 2000 \\
& $t_{\text{thres}}$ & 20 & 20 & 20 & 20\\
\cmidrule{2-6}
& \textbf{Policy Learning} & & & & \\
& Learning Rate & 5e-6 & 1e-6 & 5e-5 & 2e-5 \\
& \# Epochs & 2000 & 2000 & 2000 & 2000  \\
\bottomrule
\end{tabular}}
\label{table:hyperparameter}
\end{table*}

%% file: table/ablation_ensemble.tex
\begin{table*}
\small
\centering
\caption[]{Ensemble method ablation study}
\scalebox{1.0}{\begin{tabular}{@{}cccccc@{}}
\toprule
\textbf{Method} & 
\# of policies &
\fetchpick &
\fetchpush &
\handrotate &
\walker \\
\cmidrule{1-6}
Single policy baseline &
1 &
44.38\% $\pm$ 11.02\% &
67.97\% $\pm$ 4.81\% &
45.56\% $\pm$ 6.15\% &
4456.8 $\pm$ 1051.1 \\
\cmidrule{1-6}
\multirow{2}{*}{Split} &
3 &
45.99\% $\pm$ 11.14\% &
69.48\% $\pm$ 4.78\% &
40.14\% $\pm$ 4.30\% &
\textbf{5232.6} $\pm$ 920.1 \\

&
5 &
30.63\% $\pm$ 6.82\% &
62.98\% $\pm$ 5.24\% &
31.99\% $\pm$ 5.90\% &
4672.5 $\pm$ 747.7 \\

\cmidrule{1-6}
\multirow{2}{*}{Sample with replacement} &
3 &
\textbf{51.49}\% $\pm$ 7.90\% &
68.94\% $\pm$ 2.86\% &
49.16\% $\pm$ 3.68\% &
3750.0 $\pm$ 1342.7 \\

&
5 & 
48.21\% $\pm$ 8.09\% & 
71.09\% $\pm$ 3.91\% &
49.59\% $\pm$ 5.07\% &
4929.5 $\pm$ 640.6 \\ 

\cmidrule{1-6}
\multirow{2}{*}{Shuffle} &
3 &
50.49\% $\pm$ 9.49\% &
70.11\% $\pm$ 3.32\% &
49.34\% $\pm$ 3.88\% &
4952.1 $\pm$ 637.0 \\

&
5 & 
\textbf{51.36}\% $\pm$ 6.67\% &
\textbf{72.25}\% $\pm$ 3.20\% &
\textbf{51.03}\% $\pm$ 4.33\% &
\textbf{5170.6} $\pm$ 395.2 \\
\bottomrule
\end{tabular}}
\label{table:ensemble}
\end{table*}

%% file: fig/training_progress.tex
\begin{figure*}[!t]
    \captionsetup{position=top}
    \vspace{0.1cm}
    \centering
    {\includegraphics[width=0.9\textwidth]{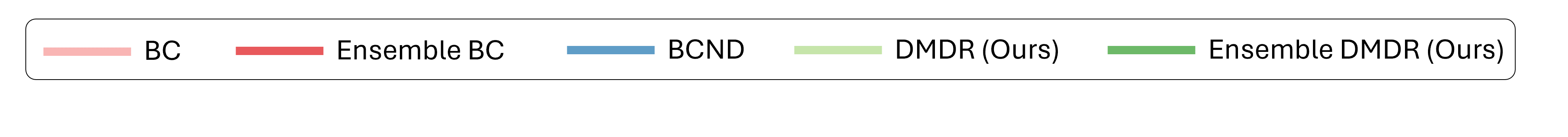}%
    \label{fig:main_legend}
    }
    \subfloat[][\fetchpick{}]
    {\includegraphics[width=0.245\textwidth]{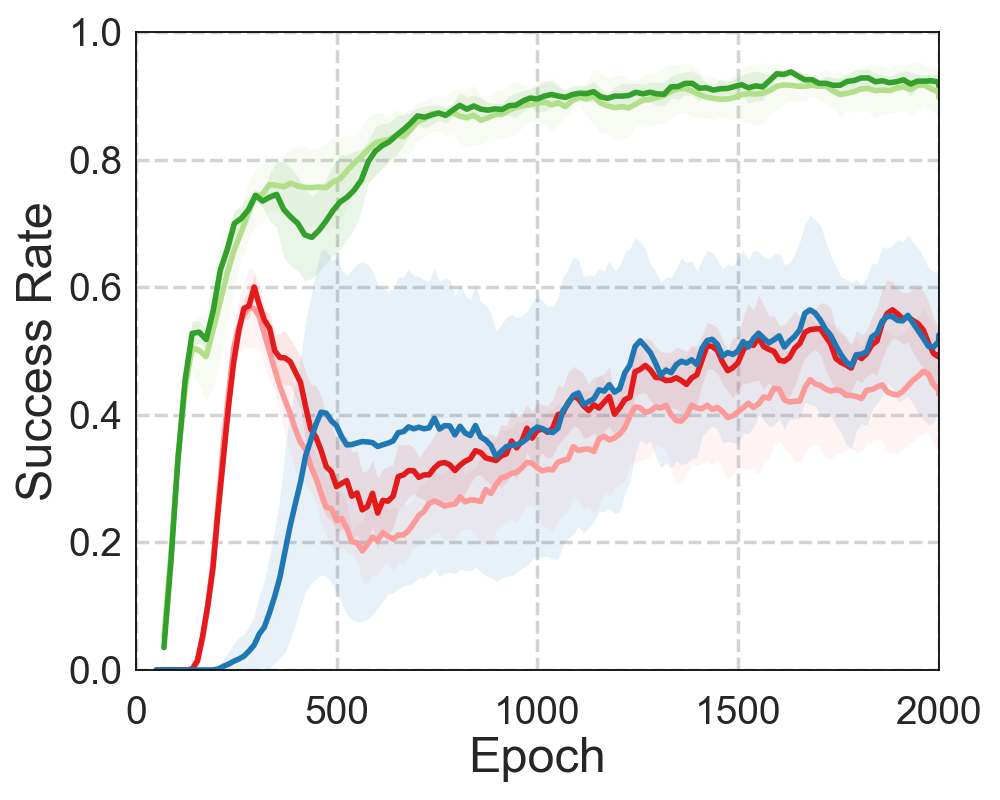}%
    \label{fig:main_pick}}
    \subfloat[][\fetchpush{}]
    {\includegraphics[width=0.245\textwidth]{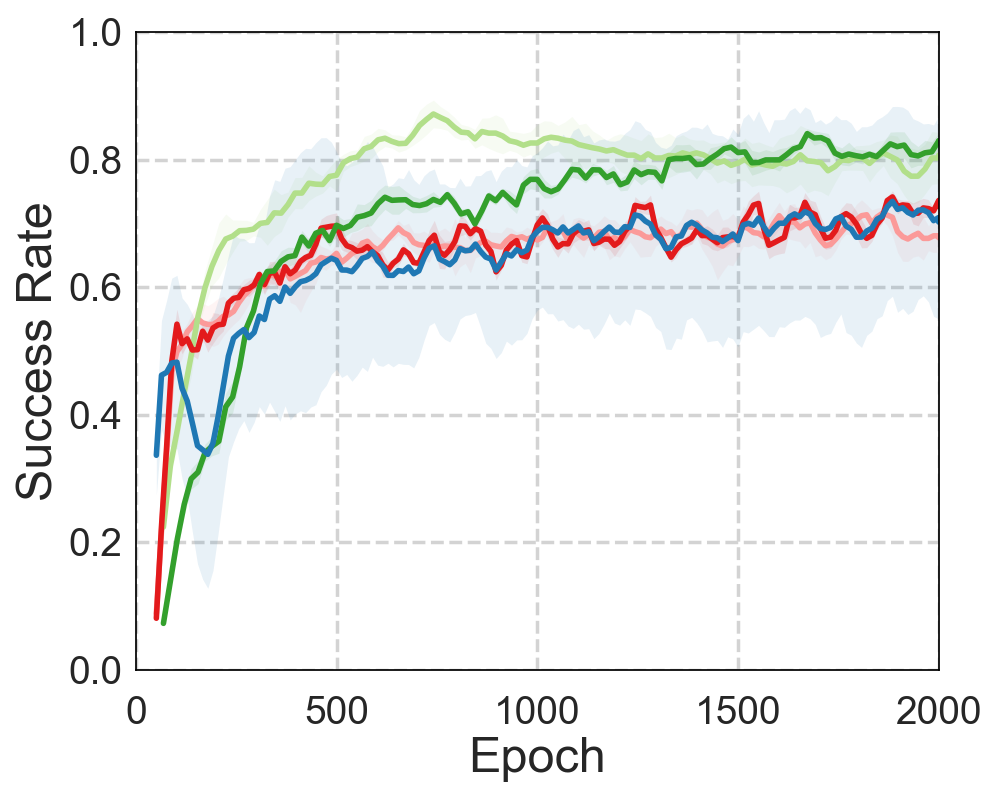}%
    \label{fig:main_push}}
    \subfloat[][\handrotate{}]
    {\includegraphics[width=0.245\textwidth]{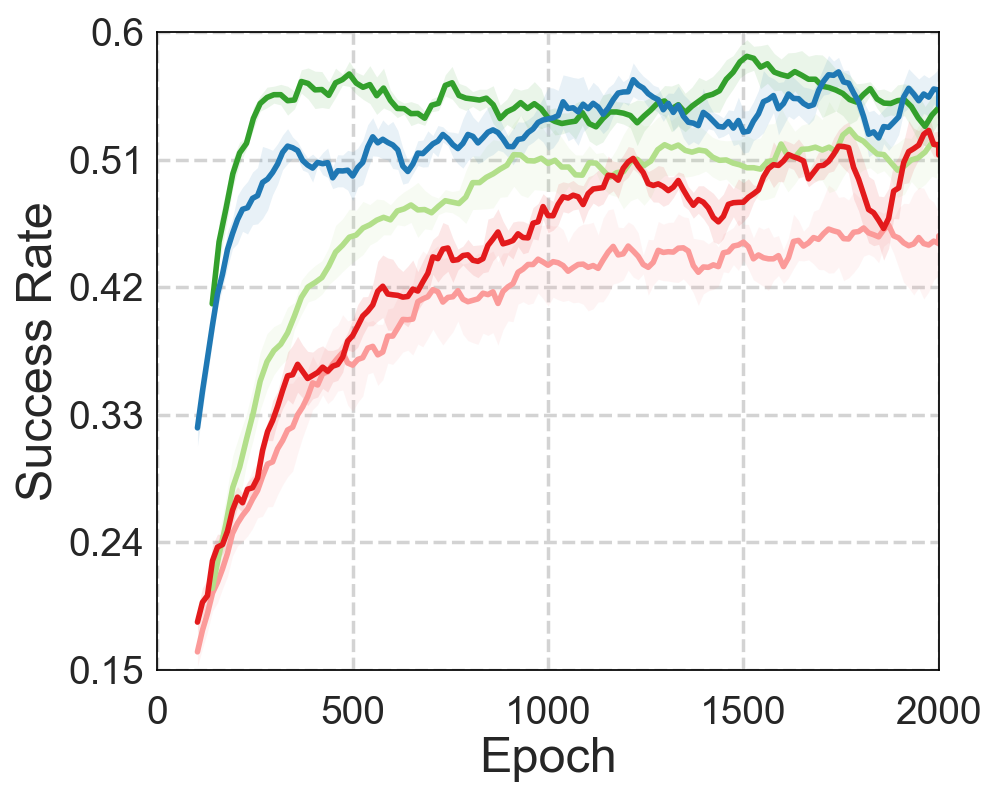}%
    \label{fig:main_hand}}
    \subfloat[][\walker{}]
    {\includegraphics[width=0.245\textwidth]{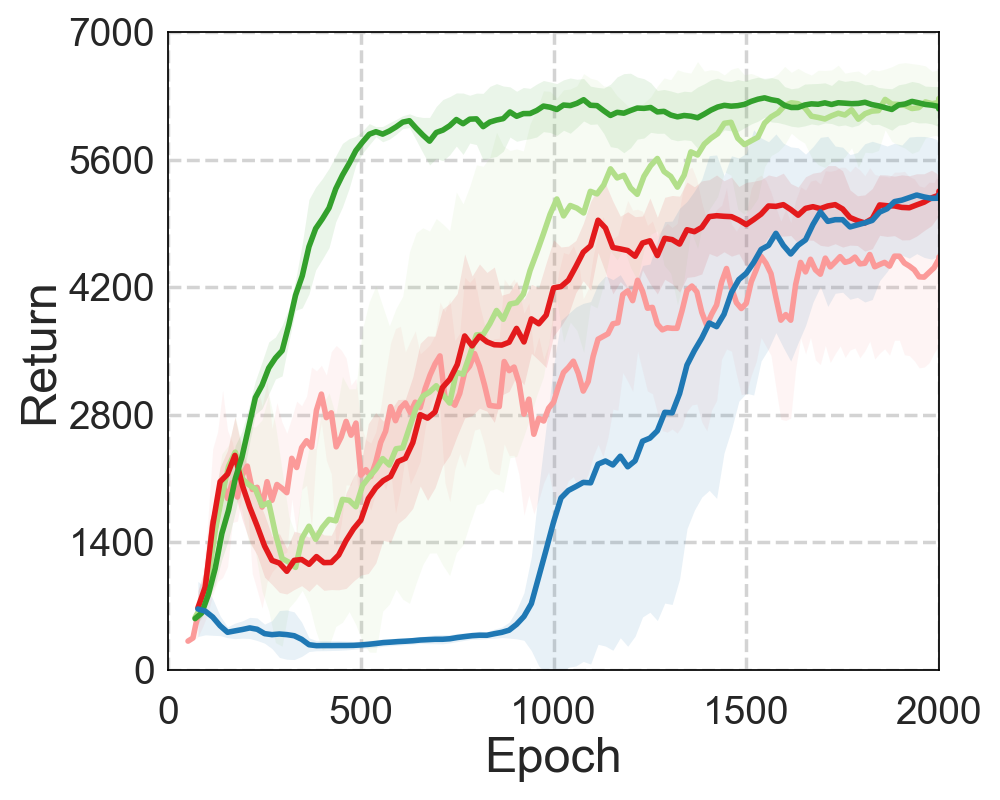}%
    \label{fig:main_walker}}    
    \caption[]{
    \textbf{Training progress.}
    We observe that \method{} is more robust and stable during the training, resulting in a lower standard deviation. In contrast, the training of BC is more easily affected by noisy samples, which can be observed on (a) \fetchpick{} and (d) \walker{}.
    }
    \label{fig:training_progress}
\end{figure*}